\documentclass[final]{siamltex}

\usepackage{subcaption}
\usepackage{hyperref} %BUG if put after

\pdfoutput=1

\usepackage[latin1]{inputenc}
\usepackage{mystyle}
\usepackage{url}

\usepackage{algorithm}
\usepackage{algorithmic}

\graphicspath{{./images/}}

\hypersetup{  
   bookmarks=true,
   backref=true,
   pagebackref=false,
   colorlinks=true,
   linkcolor=blue,
   citecolor=red,
   urlcolor=blue,
   pdftitle={A Smoothed Dual Approach for Variational Wasserstein Problems},
   pdfauthor={Marco Cuturi and Gabriel Peyr\'e},
   pdfsubject={A Smoothed Dual Approach for Variational Wasserstein Problems}
}

\title{A Smoothed Dual Approach for Variational Wasserstein Problems}

\author{Marco Cuturi and Gabriel Peyr\'e\thanks{CNRS and CEREMADE, Universit\'e Paris-Dauphine, Place du Mar\'echal De Lattre De Tassigny, 75775 PARIS CEDEX 16, FRANCE.} }

\begin{document}

\maketitle

% !TEX root = ../WassersteinDual.tex

\begin{abstract}
Variational problems that involve Wasserstein distances have been recently proposed to summarize and learn from probability measures. Despite being conceptually simple, such problems are computationally challenging because they involve minimizing over quantities (Wasserstein distances) that are themselves hard to compute. 
We show that the dual formulation of Wasserstein variational problems introduced recently by~\cite{Carlier-NumericsBarycenters} can be regularized using an entropic smoothing, which leads to smooth, differentiable, convex optimization problems that are simpler to implement and numerically more stable. We illustrate the versatility of this approach by applying it to the computation of Wasserstein barycenters and gradient flows of spacial regularization functionals.
\end{abstract}
% !TEX root = ../WassersteinDual.tex

\section{Introduction}

% Optimal transport is a well established research subject at the interplay between mathematical analysis and probability~\cite{Villani03}. Optimal transport provides a rich set of tools to define a geometry for probability measures supported on a set which is itself a metric space, among which the family of Wasserstein distances (\emph{a.k.a} earth mover's distances,~\citealt{RubTomGui00}). The goal of the present work is to define tractable optimization schemes to deal with \emph{convex variational problems} involving transportation distances, the most representative of them being the computation of the barycenter of a set of histograms under the Wasserstein metric.

% Histograms are fundamental objects popularly used in machine learning to represent complex objects as frequency vectors in the probability simplex. By defining first a set of relevant features, one can then form for each object a normalized histogram that keeps track of the frequencies of each of these features---\eg\, bags-of-words for text \cite{salton1975vector}, bags-of-visual-words for images \cite{Lowe1999,oliva2001modeling}. 

To compare two histograms in the probability simplex, information divergences---the Hellinger and $\chi_2$ distances, the Kullback-Leibler and Jensen-Shannon divergences---have the advantages of being simple and fast to compute. Optimal transport distances~\cite[\S7]{villani09}---\emph{a.k.a.} Wasserstein or earth mover's distances~\cite{RubTomGui00}---require more computational effort but are more versatile: by incorporating in their definition a metric between the bins of these histograms, they can compare sparse histograms even if their support do not overlap significantly, which can be crucial when their dimension is large. Their versatility comes, however, at a price: computing optimal transport distances requires solving a costly network flow problem, whose cost scales super-cubicly with the dimension of the considered histograms. That cost becomes even more of a drawback if one attempts to study a family of histograms using the optimal transport geometry.

Despite this computational complexity, optimal transport is becoming increasingly popular in imaging sciences and related fields, such as for instance image retrieval~\cite{RubTomGui00,Pele-iccv2009}, image interpolation~\cite{Bonneel-displacement}, computational geometry~\cite{Merigot2011,Levy3d}, color image processing~\cite{2013-Bonneel-barycenter,2014-xia-siims}, image registration~\cite{MassGenTransport} and machine learning~\cite{cuturi2013sinkhorn}.

%%%%%%%%%%%%%%%%%%%%%%%%%%%%%%%%%%%%%%
\subsection{Variational Wasserstein Problems} 

 Many learning tasks on histograms, such as averaging or clustering them, can be framed as variational problems that involve distances between pairs of histograms. These problems are easily solved when such divergences are Bregman divergences~\cite{lee1999learning,banerjee2005clustering,JeffreysCentroid-2013}, but they are far more challenging when considering instead Wasserstein distances. \cite{Carlier_wasserstein_barycenter} studied the first problem of this type, the Wasserstein barycenter problem (WBP), and showed that it is related to the multi-marginal optimal transport problem. More recently,~\cite{Solomon-ICML} proposed the Wasserstein propagation-on-graphs framework and showed that it involves a very large linear program, which can only be feasibly solved for small dimensions and families of histograms. Variational problems that involve Wasserstein distances have, however, the potential to impact a very wide range of applications. Beyond their applicability to unsupervised learning problems and their ramifications into clustering mentioned in \cite{cuturi2014fast}, they have found usage in statistics to develop population estimators~\cite{BigotBarycenter}, computer graphics to perform image modification~\cite{2014-xia-siims,2013-Bonneel-barycenter,2015-solomon-siggraph} and computer vision~\cite{ZenICPR14} to summarize complex visual signals. 

Beside the computation of barycenters, it is also possible to integrate Wasserstein distances into more general variational problems. 
For instance, optimal transport distances are used as a data fidelity to perform image denoising~\cite{Burger-JKO,Lorentz}, image segmentation~\cite{RabinPapadakisSSVM,SchnorSegmentation,SchmitzerSegmentation}, and Radon transform reconstruction~\cite{AbrahamRadon,DBLP:journals/siamsc/BenamouCCNP15}.

Our aim in this paper is to propose a computational framework that is both scalable and flexible enough to minimize energies that involve not only Wasserstein distances, but also more general functions such as regularization terms. To do so, we exploit regularization, Legendre duality and the usual toolbox of convex optimization.

%%%%%%%%%%%%%%%%%%%%%%%%%%%%%%%%%%%%%%
\subsection{Previous Works} 

\cite{cuturi2014fast} proposes to leverage the entropic regularization of Wasserstein distances introduced by \cite{cuturi2013sinkhorn} to study the WBP. Their formulation requires, however, to run a numerical subroutine, the Sinkhorn fixed-point iteration, to evaluate these objectives and compute their gradients. On the other hand, \cite{Carlier-NumericsBarycenters} show that the Fenchel-Legendre dual of the Wasserstein distance as well as its subgradients can be obtained in \emph{closed form} using nearest-neighbor assignments, that is without having to solve a single optimal transport problem. The authors do, however, struggle with non-differentiable objective functions and use a L-BFGS first order scheme. More recently, \cite{DBLP:journals/siamsc/BenamouCCNP15} have proposed an a generalized version of Sinkhorn's algorithm to compute barycenters based on Bregman's projections. This approach is useful for the barycenter problem, but cannot be easily adapted to solve more advanced variational problems. 

A typical use of such more involved variational problems is the approximation of gradient flows. As initially shown by~\cite{jordan1998variational}, it is indeed possible to approximate solutions of a large family of partial differential equations by iteratively minimizing some energy functional plus the Wasserstein distance to the previous iterates. We refer to Section~\ref{sec-grad-flows} for more details and references about these schemes. Following the method introduced in~\cite{Peyre-JKO}, one can approximate these iterations using entropic regularization. Our dual approach is crucial to be able to tackle non-separable energies, such as for instance the total variation of images.

Another illustration of the usefulness of our dual approach is the application to image segmentation developed in~\cite{RabinPapadakisSSVM}. Note that this application requires to compute the gradient of the dual of the smoothed Wasserstein distance with respect to two histograms. This formula is provided in Appendix~\ref{sub:two}.

% \todo{Explain limits by adding reference to Entropic Wasserstein Gradient Flows? we should definitely add that reference}

%%%%%%%%%%%%%%%%%%%%%%%%%%%%%%%%%%%%%%
\subsection{Contributions}

Our main contribution is to combine the strengths of the dual formulation of~\cite{Carlier-NumericsBarycenters} with the smoothing strategy laid out by~\cite{cuturi2014fast} to obtain a \emph{smooth} optimization problem whose objectives and derivatives can be computed in \emph{closed form} in \S\ref{sec:optimtransentrop}. We show that this approach can be readily used to compute Wasserstein barycenters in \S\ref{sec:dualalgo}, and explain why using regularized Wasserstein distances might be beneficial to recover smooth solutions. We proceed with more general energies that involve not only Wasserstein distances, but also more generally spatial regularization of barycenters and gradient flows, in \S\ref{sec:exten}. 

The source code to reproduce the numerical illustration of this article can be found online\footnote{\url{2015-SIIMS-wasserstein-dual}}.

%  and learn dictionaries from histogram data with the Wasserstein metric as a fitting criterion in \S\ref{sec:nmf}.

%\item Third, we show in Sections~\ref{sec:exten} that this approach is versatile and can serve as a blueprint to solve more advanced Wasserstein variational problem, such as Wasserstein propagation, as well as penalized or constrained barycenter problems. We provide a testing ground for these ideas by studying in depth the problem of averaging \emph{unnormalized measures} under a suitable extension of the Wasserstein metric. Indeed, in some application fields, the ability to take into account not only the distribution of various measures, but also their relative mass, can be crucial. We propose an algorithmic answer to this problem and illustrate it by addressing the hard problem of averaging brain activation maps defined on the folded triangulated cortex, \ie producing a mean brain activation in a group of subjects. We present results both on realistic simulations as well as on publicly available magneto-encephalography (MEG) data.

%%%%%%%%%%%%%%%%%%%%%%%%%%%%%%%%%%%%%%
\subsection{Notations}

When used on matrices, functions such as $\log$ or $\exp$ are always applied element-wise. For two matrices (or vectors) $A,B$ of the same size, $A\circ B$ (resp. $A/B$) stands for the element-wise product (resp. division) of $A$ by $B$. If $u$ is a vector, $\diag(u)$ is the diagonal matrix with diagonal $u$. $\ones_{\n} \in \RR^{\n}$ is the (column) vector of ones.

% !TEX root = ../WassersteinDual.tex

\section{Legendre Transforms of the Smoothed Wasserstein Distance}\label{sec:optimtransentrop}

We introduce in this section the entropic regularization of the Wasserstein distance, study its Legendre transform and show that it admits a simple closed form.

\subsection{Optimal Transport with Entropic Smoothing}

We consider two discrete probability distributions on the same space, represented through their histograms $\p,\q\in \Si_n$ of $\n$ values. We also introduce a symmetric cost matrix $M = (M_{ij})_{i,j=1,\ldots,n} \in \RR_+^{\n \times \n}$. Each element $M_{ij}$ accounts for the (ground) cost of moving mass from bin $i$ to bin $j$. In many applications of optimal transport, the cost matrix $M$ is defined through $n$ points $(x_i)_i$ taken in a metric space $(\Xcal,D)$ such that $M_{ij} = D(x_i,x_j)^\rho, \rho\geq 1$. Note however that we make no assumption on $M$ in this paper other than the fact that it is symmetric and non-negative. 

Given $\p,\q$, the set of couplings $U(\p,\q)$ and the discrete entropy of any coupling in that set are defined as,
\begin{equation}\begin{aligned}
	U(\p,\q) &\defeq \enscond{X \in \RR_+^{\n \times \n}}{ X \ones_\n = \p, X^T \ones_\n = \q },\; E(X) &\defeq - \sum_{ij} h(X_{ij}),
\end{aligned}
\end{equation}
where $\forall x>0, h(x) \defeq x\log x, h(0)=0$. 
We follow~\cite{cuturi2013sinkhorn}'s approach and introduce a entropy-regularized optimal transport problem:
\eql{\label{eq-entropic-wass-dist}
	W_\ga(\p,\q) \defeq \umin{X \in U(\p,\q)} \dotp{M}{X} - \ga E(X),
}
where $\ga\geq 0$. For $\ga=0$, one recovers the usual optimal transport problem, which is a linear program. $W_0$ is known as the Wasserstein distance (or Earth Mover's Distance, EMD) between $\p$ and $\q$. For $\ga>0$, Problem~\eqref{eq-entropic-wass-dist} is strongly convex and admits a unique optimal coupling $X_\ga^\star$. \cite{cuturi2013sinkhorn} called the resulting cost $\dotp{M}{X_\ga^\star}$ the \emph{Sinkhorn divergence} between $\p$ and $\q$.

While $X_\ga^\star$ is not necessarily unique for $\ga=0$, we show in the following proposition that in the small $\ga$ limit, the regularization captures the maximally entropic coupling. %(the one whose mass is maximally spread out in some sense). %When there is no risk of confusion, we note $W$ instead of $W_\ga$.
 
 \begin{proposition}
 	One has $W_\ga \rightarrow W_0$ as $\ga \rightarrow 0$, and denoting $X_\ga^\star$ the unique solution of~\eqref{eq-entropic-wass-dist}, one has
 	\eql{\label{eq-limit-problem}
 		X^\star_\ga \longrightarrow 
 		X_0^\star = \uargmax{X \in U(\p,\q)} \enscond{ E(X) }{ \dotp{M}{X} = W_0(\p,\q) }.
 	} 
 \end{proposition}
 \begin{proof}
 	We consider a sequence $(\ga_\ell)_\ell$ such that $\ga_\ell \rightarrow 0$ and $\ga_\ell > 0$.	
 	We denote $X_\ell = X^\star_{\ga_\ell}$. Since $U(\p,\q)$ is bounded, we can extract a sequence (that we do not relabel for sake of simplicity) such that $X_\ell \rightarrow X^\star$. Since $U(\p,\q)$ is closed, $X^\star \in U(\p,\q)$. We consider any $X$ such that $\dotp{M}{X} = W_0(\p,\q)$. By optimality of $X$ and $X_\ell$ for their respective optimization problems (for $\ga=0$ and $\ga=\ga_\ell$), one has
 	\eql{\label{eq-proof-gamma-conv}
 		0 \leq \dotp{M}{X_\ell} - \dotp{M}{X} \leq \ga_\ell ( E(X_\ell)-E(X) ).
 	}
 	Since $E$ is continuous, taking the limit $\ell \rightarrow +\infty$ in this expression shows that 
 	$\dotp{M}{X^\star} = \dotp{M}{X}$ so that $X^\star$ is a feasible point of~\eqref{eq-limit-problem}. Furthermore, dividing by $\ga_\ell$ in~\eqref{eq-proof-gamma-conv} and taking the limit shows that 
 	$E(X) \leq E(X^\star)$, which shows that $X^\star$ is a solution of the maximization~\eqref{eq-limit-problem}. Since the solution $X_0^\star$ to this program is unique by strict convexity of $-E$, one has $X^\star = X_0^\star$, and the whole sequence is converging. 
 \end{proof}

\cite{cuturi2014fast} provided a dual expression for $W_\ga$. The proof of that result follows from an application of Fenchel-Rockafellar duality to the primal problem~\eqref{eq-entropic-wass-dist}. The indicator function of a closed convex set $\Cc$ is $\iota_\Cc(x)=0$ for $x \in \Cc$ and $\iota_\Cc(x)=+\infty$ otherwise.

\begin{proposition}
	One has 
	\eql{\label{eq-wassdist-dual}
		W_{\gamma}(\p,\q) = \umax{u,v\in\RR^n}  \dotp{u}{p} + \dotp{v}{q} - B( u,v ),
	}
	\eq{B( u,v ) \eqdef
		\begin{cases} 
			\gamma \sum_{i,j} \exp(\frac{1}{\gamma}(u_i + v_j - M_{ij})-1),\text{ if } \gamma>0;\\	 				\iota_{\Cc_M}(u,v), \text{ if } \gamma=0, \quad\text{where}\quad \Cc_M \eqdef 
				\enscond{(u,v)}{ u_i + v_j \leq M_{ij}}.
		\end{cases}
	}
\end{proposition}

When $\ga>0$, this regularization results in a smoothed approximation of the Wasserstein distance with respect to either of its arguments, as shown below. To simplify notations, let us introduce the notation $H_\q(\p)$, the Wasserstein distance of any point $\p$ to a fixed histogram $\q\in\Sigma_n$,
\eq{
	\foralls p \in \Si_n, \quad H_q(p) \defeq  W_\ga(\p,\q).
} 
Note that $H_q$ is a convex function for all $\ga\geq 0$. When $\ga>0$, $H_q$ has the following properties, which follow from the direct differentiation of expression~\eqref{eq-wassdist-dual}:
\begin{proposition}\label{prop-primal-properties} For $\ga>0$ and $(\p,\q) \in \Si_n \times \Si_n$ with $p>0, q>0$, 
	 $H_q$ is $C^1$ at $p$ and $\nabla H_q(p) = u^\star$ where $u^\star$ is the unique solution of~\eqref{eq-wassdist-dual} satisfying $\dotp{u^\star}{\ones_n}=0$.
\end{proposition}

Computing both $H_q$ and its gradient requires thus the resolution of the optimization problem in Eq.~\eqref{eq-wassdist-dual}, which can be solved with a Sinkhorn fixed-point iteration~\cite{cuturi2013sinkhorn} as remarked by \cite[\S5]{cuturi2014fast}. This computation can be avoided when studying the Fenchel-Legendre conjugate of $H_q$, as shown below.

\subsection{Legendre Transform with Respect to One Histogram}

The goal of this section is to show that the Fenchel-Legendre transform of $H_q$,
\eq{
	\foralls g \in \RR^n, \quad
	H_q^*(g) = \umax{p \in \Si_n} \dotp{g}{p} - H_q(p),
}
has a closed form. This result was already known when $\ga=0$, that is for the original Wasserstein distance. \cite[Prop. 4.1]{Carlier-NumericsBarycenters} showed indeed that computing $H_q^*$ only requires a sequence of nearest-neighbor assignments. We show that for $\ga>0$, these nearest-neighbor assignments are replaced by soft assignments.

% The main result of this paper is to show that the Legendre transform of the \emph{smoothed} Wasserstein distance ($\ga>0$) has a closed form. 
Compared to the primal smoothed Wasserstein distance $H_q$, the computation of both $H_q^*$ and its derivatives can be carried out without having to solve a matrix-scaling problem. These properties are at the core of the computational framework we develop in this paper.

\begin{theorem}[Legendre Transform of $H_q$]\label{thm-legendre-transf}
	For $\ga>0$, the Fenchel-Legendre dual function $H_q^*$ is $C^\infty$. Its gradient function $\nabla H_q^*(\cdot)$ is $1/\gamma$ Lipschitz. Its value, gradient and Hessian at $g\in\RR^n$ are, writing $\al=e^{g/\ga}$ and $K=e^{-M/\ga}$,
	\begin{equation}\label{eq-obj-dual}\begin{aligned}
		H_q^*(g) &= \ga \left( E(q)+\dotp{q}{\log K\alpha}\right),\, \nabla H_q^*(g) = \al \circ \pa{\K  \frac{q}{\K  \al} } \in \Sigma_n,\\
		\nabla^2 H_q^*(g)  &= \frac{1}{\ga}\diag\left(\al \circ K \frac{q}{\K \al}\right) - \frac{1}{\ga}\diag(\al) K \diag\left(\frac{q}{(\K  \al)^2 }\right)\K  \diag(\al).\end{aligned}
\end{equation}
%	where $\circ$ and $/$ denote respectively element-wise multiplication and division, and $e^{A} = (e^{A_{ij}})_{ij}$.

\end{theorem}

\begin{proof}
%	The fact that $H_{q}$ is $\ga$ strongly convex is proved in~\cite{CuturiSinkhorn} \todo{not really}, and this implies the smoothness of $H_q^*$, see for instance~\cite{}. 
	Writing $H_{q,M}(p)$ in place of $H_{q}(p)$ to make explicit the dependency on $M$, one has
	\begin{align*}
		H_{q,M}^*(g) &= \umax{p\in\Si_{\n}} \dotp{g}{p} - \umax{u,v} \dotp{u}{p} + \dotp{v}{q} - B( u,v ) \\
	%	&= \umax{p} - \umax{u,v} \dotp{u-g}{p} + \dotp{v}{q} + B( u,v ) \\
		&= \umax{p\in\Si_{\n}} - \umax{u',v} \dotp{u'}{p} + \dotp{v}{q} - B( u'+g,v) \\
%		&= \umax{p\in\Si_{\n}} - \umax{u,v} \dotp{u}{p} + \dotp{v}{q} - \beta_{\ga,M-g\ones ^T}( u,v ) %\\&
		&= \umax{p\in\Si_{\n}} - H_{q,M-g\ones^T}(p) = -\umin{p\in\Si_{\n}} \umin{X \in U(\p,\q) } \dotp{M-g\ones^T}{X} - \ga E(X).	
	\end{align*}
	This leads to an optimal transport problem which is only constrained by \emph{one} marginal,
	\eql{\label{eq-grad-dual-proof-1}
		H_{q,M}^*(g) = -\!\!\!\!\!\umin{X, X^T\ones  = q, X\geq 0} \dotp{M-g\ones^T}{X} - \ga E(X)\\				
	}
which can be explicitly solved by writing first order conditions for~\eqref{eq-grad-dual-proof-1} to obtain that, at the optimum, we necessarily have $\log(X_{ij}^\star)=\frac{1}{\gamma}(g_i-M_{ij}-\rho_j)-1$ for some vector of values $\rho\in\RR^n$. Therefore $X^\star$ has the form $X^\star=\diag(\alpha)K \diag(e^{\rho/\gamma-1})$, using the notation $\al=e^{g/\ga}$. Because of the marginal constraint that $X^{\star T}\ones =q$, the rightmost diagonal matrix must necessarily be equal to $\diag(q /\K \alpha)$, and thus 
	$X^\star=\diag(\alpha)K \diag(q/\K \alpha).$
Therefore, the Legendre transform $H_{q,M}^*$ has a closed form,
	\begin{equation}\label{eq:originalexpression}H_{q,M}^*(g) = -\dotp{M-g\ones^T}{X^\star}+\gamma E(X^\star)\end{equation} 
	which can be simplified to 
	$$H_{q,M}^*(g) =	-\ga \ones_d^T \left((K\alpha) \circ h(q/K\alpha)\right).$$
	by using the fact that $X^\star=\diag(\alpha)K \diag(q/\K \alpha)$. This equation can be simplified further to obtain the expression provided in Eq.~\eqref{eq-obj-dual}. Using Eq.~\eqref{eq:originalexpression}, we have that
	\eq{
		\nabla H_{q,M}^*(g) = X^\star \ones  = \al \circ \pa{K  \frac{q}{\K  \al} }.
	}
	Computations for the Hessian follow directly, and result in the expression given Eq.~\eqref{eq-obj-dual}. Since the Hessian can be written as the difference of two positive definite matrices, one diagonal and the other equal to the product of a matrix times its transpose, the trace of $\nabla^2 H_q^*(g)$ is upper bounded by the trace of the first term, which is equal to $\frac{1}{\gamma}$ (recall that $\nabla H_{q,M}^*(g)$ is in the simplex), which proves the $\frac{1}{\gamma}$-Lipschitz continuity of the gradient of $H_q^*$.
\end{proof}

In some settings, such as the Wasserstein propagation framework of~\cite{Solomon-ICML}, the aim is to minimize Wasserstein distances with respect to two variable arguments. We provide the formulation for the corresponding Legendre transform in Theorem~\ref{eq-obj-bothvar} in the Appendix.

%%%
\subsection{Un-regularized Case}

The result of Theorem~\ref{thm-legendre-transf} is derived in the un-regularized case (\ie\, $\ga=0$) in~\cite{Carlier-NumericsBarycenters}. For the sake of comparison, let us now recall this result using our notations. Given a cost matrix $M\in\RR^{n\times n}$ and a vector $g\in\RR^n$, we introduce for $i\leq n$ the set $N_{M,g}(i) = \argmin_k M_{ik}-g_i.$
 In other words, $N_{M,g}(i)$ is the set of nearest-neighbors of $i$ with respect to the vector of distances $M_{ik}$ offset by $-g_i$. 
 
A map $\sigma_{M,g}: \{1,\dots,n\}\rightarrow \Sigma_n$ is called a nearest-neighbor map if the vector $\sigma_{M,g}(i)$ only has non-zero values on indices in $N_{M,g}(i)$, namely 
$$ 
	[\sigma_{M,g}(i)]_j \ne 0 \quad\Longleftrightarrow\quad j\in N_{M,g}(i).
$$
If $N_{M,g}(i)$ is a singleton $\{j\}$ (the minimization $\min_k M_{ik}-g_i$ admits only one optimal solution) then $\sigma_{M,g}(i)$ is necessarily equal to a Dirac histogram $\delta_j$ (we call a Dirac histogram a histogram with mass $1$ on only one coordinate, of index $j$ in this case). When $N_{M,g}(i)$ has more than one element, ties have to be taken care off, and this can be carried out arbitrarily, for instance by dividing the mass equally among those nearest neighbors, or by only choosing arbitrarily one of them. We can now recall the result of \cite{Carlier-NumericsBarycenters}:
 
\begin{proposition}[Carlier et al. 2014, Prop. 4.1]\label{prop:harddual} For $\ga=0$ and a nearest-neighbor map $\sigma_{M,g}$, the Fenchel-Legendre dual function $H^*_q$ admits the following vector in its sub-differential $\partial H_{\q}^*(g)$ at $g\in\RR^n$,		
$$ 
	S_q(g) \defeq \sum_{i\leq d}q_i\sigma_{M,g}(i) \in \partial H_{q}^*(g).
$$ 
Note that $S_q(g)$ is in $\Sigma_n$. The value of $H_{\q}^*(g)$ is $\dotp{S_q(g)}{g}$.
\end{proposition}

% !TEX root = ../WassersteinDual.tex

\section{Smooth Dual Algorithms For the Wasserstein Barycenter Problem}\label{sec:dualalgo}

In this section, we use the properties of the Legendre transform of the Wasserstein distance as detailed in Section~\S\ref{sec:optimtransentrop} to solve the Wasserstein Barycenter Problem.

%%%%%%%%%%%%%%%%%%%%%%%%%%%%%%%%%%%%%%%%%%%%%%%%%%%%%%%%%%%%%
\subsection{Smooth Dual Formulation of the WBP}

Following the introduction of the Wasserstein Barycenter Problem (WBP) by~\cite{Carlier_wasserstein_barycenter},~\cite{cuturi2014fast} introduced the smoothed WBP with $\ga$-entropic regularization ($\ga$-sWBP) as
\eql{\label{eq-variational-barycenter-discrete}
	\umin{p \in \Si_n}
		\sum_{k=1}^N \la_k  H_{\q_k}(\p) \enspace .
}
where $(\q_1,\ldots,\q_N)$ is a family of histograms in $\Si_n$. When $\ga=0$, the $\ga$-sWBP is exactly the WBP. In that case, problem \eqref{eq-variational-barycenter-discrete} is in fact a linear program, as discussed later in \S\ref{sec:wassbarprob}. When $\ga>0$ the $\ga$-sWBP is a \emph{strictly} convex optimization problem that admits a unique solution, which can be solved with a simple gradient descent as advocated by \cite{cuturi2014fast}. They show that the $N$ gradients $\left[\nabla H_{\q_k}(\p) \right]_{k \leq \N}$ can be computed at each iteration by solving $\N$ Sinkhorn matrix-scaling problems. Because these gradients are themselves the result of a numerical optimization procedure, the problem of choosing an adequate threshold to obtain sufficiently precise gradients arises as a key parameter in that approach.
We take here a different route to solve the $\ga$-sWBP, which can be either interpreted as a smooth alternative to the dual WBP studied by~\cite{Carlier-NumericsBarycenters}, or the dual counterpart to the smoothed WBP of \cite{cuturi2014fast}.

\begin{theorem}\label{prop-dual-energy}
	The barycenter $p^\star$ solving~\eqref{eq-variational-barycenter-discrete} satisfies 
	\eql{\label{eq-primal-dual-relationship}
		\foralls k = 1,\ldots,\N, \quad 
		p^\star = \nabla H_{\q_k}^*(g_k^\star)
	}
	where $( g_k^\star )_k$ are any solution of the smoothed dual WBP:
	\eql{\label{eq-dual-pbm}
		\umin{ g_1,\ldots, g_N\in\RR^n} \sum_k \la_k H_{\q_k}^*(g_k)
		\qstq \sum_k \la_k g_k = 0.
	}
\end{theorem}

\begin{proof}
	We re-write the barycenter problem 
	$$\umin{\p_1,\ldots,\p_N} \sum_k \la_k H_{\q_k}(\p_k) \qstq \p_1=\ldots=\p_N$$
	whose Fenchel-Rockafelar dual reads
		$$\umin{\tilde g_1,\ldots,\tilde g_N} \sum_k \la_k H_{\q_k}^*(\tilde g_k/\la_k) 
		\qstq \sum_k \tilde g_k = 0.$$
	Since the primal problem is strictly convex, the primal-dual relationships show that the unique solution $p^\star$ of the primal can be obtained from any solution $(g_k^\star)_k$ via the relation
	$\p_k^\star = \nabla H_{\q_k^\star}^*(\tilde g_k^\star/\la_k)$.
	One obtains the desired formulation using the change of variable $g_k = \tilde g_k/\la_k$.
\end{proof}

Theorem~\ref{prop-dual-energy} provides a simple approach to solve the $\ga$-sWBP: Rather than minimizing directly the sum of regularized Wasserstein distances in Eq. \eqref{eq-variational-barycenter-discrete}, this formulation only involves minimizing a strictly convex function with closed form objectives and gradients. 

\paragraph{Parallel Implementation} The objectives, gradients and Hessians of the Fenchel-Legendre dual $H^*_q$ can be computed using either matrix-vector products or element-wise operations. Given $N$ histograms $(q_k)_k$, $N$ dual variables $(g_k)_k$ and $N$ arbitrary vectors $(x_k)_k$, the computation of $N$ objective values $(H_{q_k}^*(g_k))_k$ and $N$ gradients $(\nabla H_{q_k}^*(g_k))_k$ can all be vectorized. Assuming that all column vectors $g_k$, $q_k$ and $x_k$ are gathered in $n\times N$ matrices $G$, $Q$ and $X$ respectively, we define first the following $n\times N$ auxiliary matrices:
$$ A\defeq e^{G/\ga}, \quad B\defeq K A, \quad C\defeq \frac{Q}{B} , \quad  \Delta\defeq A\circ (KC),$$
to form the vector of objectives
$$\label{eq:onehistoobj}H^*\; \defeq [ H_{q_1}^*(g_1),\dots, H_{q_N}^*(g_N)]= -\ga \ones_n^T \left(Q \circ \log(C)\right),$$
and the matrix of gradients 
\begin{equation}\label{eq:onehistograd}\nabla H^*\; \defeq [\nabla H_{q_1}^*(g_1),\dots,\nabla H_{q_N}^*(g_N)]=\Delta.\end{equation}
% \begin{equation}\label{eq:onehistohessian}\begin{aligned}\nabla^2 H^* X \; &\defeq \left[ \nabla^2 H_{q_1}^*(g_1) x_1,\dots, \nabla^2 H_{q_N}^*(g_N) x_N\right] \\&=\frac{1}{\gamma} \left[\Delta\circ X - A \circ \left(K  \left(\frac{Q\circ \left(K \left(A\circ X\right)\right)}{B\circ B} \right)\right) \right].\end{aligned}\end{equation}
%\begin{equation}\label{eq:onehistohessian}\begin{aligned}\nabla^2 H^* X =\frac{1}{\gamma} \left[\Delta\!\circ\! X - A\!\circ\! \left(K  \left(\frac{Q\!\circ\!\left(K\left(A\!\circ\!X\right)\right)}{B\circ B} \right)\right) \right].\end{aligned}\end{equation}

%%%%%%%%%%%%%%%%%%%%%%%%%%%%%%%%%%%%%%%%%%

% 
% \subsection{Parallel Computations, Two Histograms} 
% 
% We precompute the denominators appearing in Eq.~\eqref{eq:xgh} for all pairs $(g_k,h_k)_{k\leq N}$ using a $n\times n$ matrix times $n\times N$ matrix of variables, a Schur product and a simple matrix vector product: 
% $$U=\ones_n^T \left(e^{G/\ga}\circ (Ke^{H/\ga})\right).$$
% $U$ is therefore a $1 \times N$ row vector.
% The $2n\times N$ matrix of gradients is then obtained as
% $$\nabla W^*\;\defeq [\nabla W^*(g_1,h_1),\dots,\nabla W^*(g_N,h_N)]= \begin{bmatrix}e^{G/\ga} \circ (K e^{H/\ga}) \\ e^{H/\ga} \circ (\K  e^{G/\ga})\end{bmatrix}\circ \left(\ones_{2n} 1/U \right),$$
% where the product by a diagonal matrix above should be implemented in practice using a fast subroutine such as \texttt{bsxfun} in Matlab, broadcasting with Numpy in Python or any convenient way to scale rows in a matrix.

%%%%%%%%%%%%%%%%%%%%%%%%%%%%%%%%%%%%%%%%%%%%%%%%%%%%%%%%%%%%%
\subsection{Algorithm}
The $\ga$-sWBP in Eq.~\eqref{eq-dual-pbm} has a smooth objective with respect to each of its variables $g_k$, a simple linear equality constraint, and both gradients and Hessians that can computed in closed form. We can thus compute a minimizer for that problem using a naive gradient descent outlined in Algorithm~\ref{algo:firstorder}. To obtain a faster convergence, it is also possible to use accelerated gradient descent, quasi-Newton or truncated Newton methods~\cite[\S10]{Boyd:1072}. In the latter case, the resulting KKT linear system is sparse, and solving it with preconjugate gradient techniques can be efficiently carried out. We omit these details and only report results using off-the-shelf L-BFGS. From the dual iterates $g_k$ stored in a $n\times N$ matrix $G$, one recovers primal iterates using the formula~\eqref{eq-primal-dual-relationship}, namely $\p_k = e^{g_k/\ga}\circ K \frac{q_k}{Ke^{g_k/\ga}}.$
At each intermediary iteration one can thus form a solution to the smoothed Wasserstein barycenter problem by averaging these primal solutions,$\tilde{\p} = \Delta\ones_N/N.$ Upon convergence, these $\p_k$ are all equal to the unique solution $\p^\star$. The average at each iteration $\tilde{\p}$ converges towards that unique solution, and we use the sum of all line wise standard deviations of $\Delta$: $\ones_d^T\sqrt{(\tilde{\Delta}\circ \tilde{\Delta}) \ones_N/N}$ where $\tilde{\Delta}=\Delta(I_N-\frac{1}{N}\ones_N\ones_N^T)$ to monitor that convergence in our algorithms.

\begin{algorithm}
	\begin{algorithmic}[1]
		\caption{Smoothed Wasserstein Barycenter, Generic Algorithm\label{algo:firstorder}}
		\STATE \textbf{Input}: $Q=[\q_1,\cdots,\q_N] \in(\Sigma_n)^N$, metric $M\in\RR_+^{n\times n}$, barycenter weights $\lambda\in\Sigma_N$, $\gamma>0$, tolerance $\varepsilon>0$. 
		\STATE initialize $G\in\mathbb{R}^{n\times N}$ and form the $n\times n$ matrix $K=e^{-M/\gamma }$.
		\REPEAT
				\STATE From gradient matrix $\Delta$ (see Eq.~\ref{eq:onehistograd}) produce update matrix $\hat{\Delta}$ using either $\Delta$ directly or other methods such as L-BFGS.
				\STATE $G= G - \tau \hat{\Delta}$, update with fixed step length $\tau$ or approximate line search to set $\tau$.
				\STATE $G = G - \frac{1}{\norm{\lambda}_2^2}(G \lambda)  \lambda^T$ \quad (projection such that $G\lambda=0$)
		\UNTIL{$\ones_d^T\sqrt{(\tilde{\Delta}\circ \tilde{\Delta}) \ones_N/N}<\varepsilon$, where $\tilde{\Delta}=\Delta(I_N-\frac{1}{N}\ones_N\ones_N^T)$}
		\STATE output barycenter $\p=\Delta \ones_N/N$.
	\end{algorithmic}
\end{algorithm}

%%%%%%%%%%%%%%%%%%%%%%%%%%%%%%%%%%%%%%%%%%%%%%%%%%%%%%%%%%%%%%%%%
\subsection{Initialization Heuristic}
\label{sec:magictrick}

\def\abstildek{\abs{\kappa}}
Definition~\ref{def:SI} provides an initialization heuristic to initialize both the primal and dual smoothed WBP, motivated by the fact that they provide directly the optimal primal/dual solutions when the histograms are Dirac histograms as proved in Proposition~\ref{prop:init}.

\begin{definition}[Primal and Dual WBP Initialization]\label{def:SI}
Let $(q_1,\cdots,q_N)$ be $N$ target histograms in the simplex $\Sigma_n$ and $\lambda$ a vector of weights in $\Sigma_N$. Let  $\bar{q}=\sum_k \lambda_k q_k\in\Sigma_n$. Define $\kappa_\ga$ as
$$\kappa_\gamma = \begin{cases} e^{-M\bar{q}/\ga}/(\ones_\n^Te^{-M\bar{q}/\ga}) \text{ if } \ga>0,\\  
\delta_j, \text{ where } j\in\argmin_\ell [M\bar{q}]_\ell, \text{ if } \gamma=0.\end{cases}$$ 
For $\ga\geq 0$, the $\ga$-smoothed primal and dual WBP can be initialized respectively with the following primal and $N$ dual feasible solutions:
\begin{equation}\label{eq:sinkprimalinit}
	\p^{(0)}\defeq \kappa_\ga,
\end{equation}
and for $1 \leq k \leq N$,
\begin{equation}\label{eq:sinkdualinit}
g_k^{(0)} \defeq M (q_k - \bar{q}).
\end{equation}
\end{definition}
%g_k^{(0)} \defeq = \ga \log \left(\frac{1}{\abstildek}\frac{\kappa}{k_k}\right) + \frac{\ga}{N\lambda_k}\log \left(\abstildek\right) \ones_n.

The primal initialization described above differs when $\ga>0$ or $\ga=0$: For $\ga>0$, $\kappa_\ga$ is the normalized, weighted geometric average of the columns of the kernel $K=e^{-M/\ga}$; when $\ga=0$, $\kappa_\ga$ is a vector of zero values except for a value of $1$ on the index corresponding to the (or any, if many) smallest entry of $M\bar{q}$. On the other hand, the dual initialization is the same for both smoothed and non-smoothed Wasserstein barycenter problems.

The initializations proposed in Definition~\ref{def:SI} solve the WBP in the case that all histograms are Dirac histograms, as proved in Proposition~\ref{prop:init}. For more general problems, we have observed that this initialization is particularly useful when solving the WBP with the dual formulation, but not so much with the primal one. In many experimental problems we have considered, the dual initialization seems to capture important features of the optimal solution. The primal solution that results from this dual initialization, that obtained by averaging the gradients $\nabla H_{\q_k}^*(g_k^\star)$ as suggested by the primal/dual relation of Equation~\eqref{eq-variational-barycenter-discrete}, can serve as a rough approximation of the barycenter. We provide its explicit expression $p^{(0)}_{\text{dual}}$ below. Note that $p^{(0)}_{\text{dual}}$ differs from the initialization $p^{(0)}$ suggested in Equation~\eqref{eq:sinkprimalinit}.

$$p^{(0)}_{\text{dual}} = \frac{1}{n} \left(e^{M(Q-\frac{1}{n}Q\ones_n\ones_N^T)/\ga}\circ \left(K\frac{Q}{Ke^{M(Q-\frac{1}{n}Q\ones_n\ones_N^T)}}\right)\right)\ones_n.$$

\begin{proposition}\label{prop:init} Let $\lambda$ be a vector of weights in $\Sigma_N$, and $(q_1,\cdots,q_N)$ be $N$ Dirac histograms, namely histograms that are zero everywhere but for one coordinate equal to 1. For $\ga\geq 0$, the $\ga$-sWBP primal and dual problems are solved exactly using the initialization of Definition~\eqref{def:SI}.
\end{proposition}

\begin{proof}	
To simplify notations, we write $p=p^{(0)}$ and $g_k=g_k^{(0)}$ as defined in Definition~\ref{def:SI} above. First, one can easily check that both initialization satisfy the necessary constraints, \ie~$p\in\Sigma_n$ and $\sum_k \lambda_k g_k=0$.

When $\ga=0$, since all $q_k$ are Dirac histograms, the Wasserstein distance of any point $x$ in the simplex to any $q_k$ is equal to $x^T M q_k$. Therefore, the Wasserstein barycenter objective evaluated at $x$ is equal to $x^T M \bar{q}$. This can be trivially minimized by selecting any histogram giving a mass of $1$ to the index corresponding to any smallest entry in the vector $M \bar{q}$, which is the definition of $p$. A similar computation for the dual problem results in the dual optimal outlined above.

When $\ga>0$, we need to prove that each gradient of $H^*_{\q_k}$ computed at $g_k$ is equal to $p$ for all $1 \leq k \leq \N$. Writing $\alpha_k=e^{g_k/\gamma}$, we recover that 
$$\alpha_k = \frac{\kappa_\ga}{\xi_k},$$ 
where $\xi_k\defeq e^{-M q_k/\ga}$. Since $q_k$ is a Dirac histogram, all of its coordinates are equal to $0$, but for one coordinate whose value is $1$. Let $j$ be the index of that coordinate. Therefore, $\xi_k\defeq e^{-M q_k/\ga}=K_j$, where $K_j$ is the $j^{\text{th}}$ column of the matrix $K=e^{-M/\ga}$. Therefore,
$$\alpha_k = \frac{\kappa_\ga}{K_j}.$$
Let us now compute the gradient $\nabla_k$ of $H^*_{\q_k}$ at $g_k$ by following  Eq.~\eqref{eq-obj-dual}:
$$\nabla_k = \alpha_k\circ \left(K \frac{q_k}{K\alpha_k}\right).$$
Because of the symmetry of $K$, we have that the $j^{\text{th}}$ element of the vector $K\alpha_k$ is equal to:
$$ (K\alpha_k)_j = K_j^T \alpha_k= \ones_\n^T \left( K_j \circ \alpha_k\right)= \ones_\n^T \left( K_j \circ \left(\frac{\kappa_\ga}{K_j}\right)\right)= 1.$$ 
Since only the $j^{\text{th}}$ element of $q_k$ is non-zero by definition, $q_k/(K\alpha_k)=q_k$.
Because $q_k$ is everywhere zero except for its $j^{\text{th}}$ coordinate, $K(q_k/K\alpha_k)$ is thus equal to the $j^\text{th}$ column of $K$, namely 
$$K\frac{q_k}{K\alpha_k}=K_j.$$
Finally, we obtain that the gradient of $H^*_{\q_k}$ at $g_k$ is equal to 
$$\nabla_k = \alpha_k\circ \left(K \frac{q_k}{K\alpha_k}\right)=  \frac{\kappa_\ga}{K_j} \circ K_j =\kappa_\ga=p^{(0)},$$
which holds for all indices $1\leq k\leq N$.
\end{proof}

% !TEX root = ../WassersteinDual.tex
% better to integrate it there?
%\section{Numerical Experiments}\label{sec:numericalresults}

%%%%%%%%%%%%%%%%%%%%%%%%%%%%%%%%%%%%%%%%%%%%
\subsection{Smoothing and Stabilization of the WBP}
\label{sec:wassbarprob}
We make the claim in this section that smoothing the WBP is not only beneficial computationally, but may also yield more stable computations. Of central importance in this discussion is the fact that the WBP can be cast as a LP of $Nn^2+n$ variables and $2Nn$ constraints, and thus solved \emph{exactly} for small $n$ and $N$:
\begin{equation}\begin{aligned}\label{eq-WBP-simplex}
	\min_{X_1,\cdots,X_N,p}& \sum_{k=1}^N \la_k  \dotp{X_k}{M}\\
	\text{s.t. } & X_k \in \RR^{n\times n}_+, \forall k\leq N; p\in\Sigma_n,\\
	& X_k^T \ones_n = \q_k, \forall k\leq N,\\
	 & X_1 \ones_n = \dots = X_N \ones_n= p.
\end{aligned}\end{equation}
Given couplings $X_1^\star,\cdots,X_N^\star$ which are optimal solutions to Eq.\eqref{eq-WBP-simplex}, the solution to the WBP is equal to the marginal common to all those couplings: $p^\star=X_k^\star \ones_n$ for any $k\leq N$. For small $N$ and $n$, this problem is tractable, but it can be surprisingly ill-posed as we see next.

\begin{figure}[ht]
	\centering\includegraphics[width=.75\columnwidth]{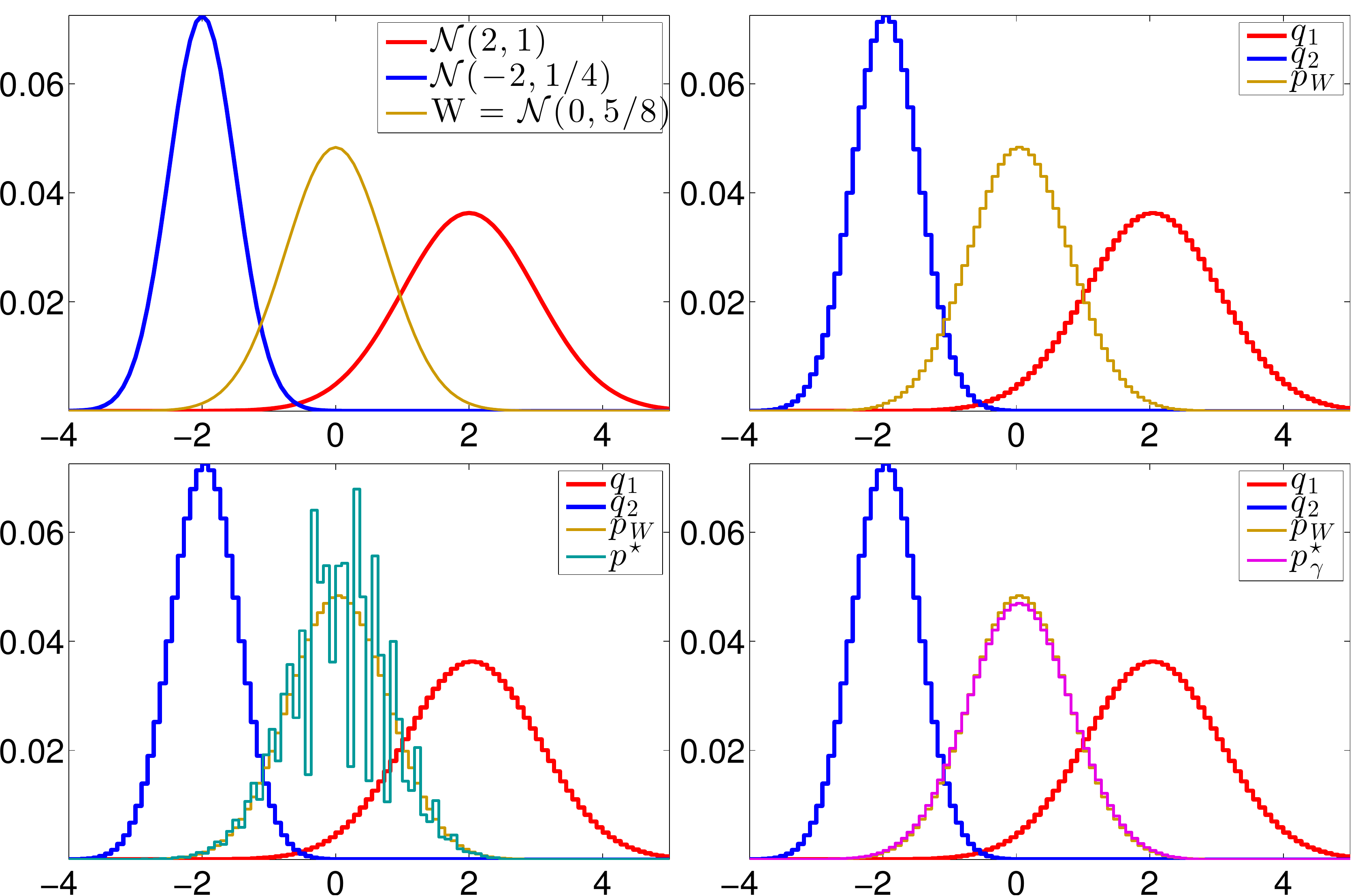}    
	\caption{(top-left) two Gaussian densities and their barycenter (top right) same densities, discretized (bottom left) discretization of the true barycenter \emph{vs.} the optimum of Equation~\ref{eq-WBP-simplex} (bottom right) barycenter computed with our smoothing approach. }\label{fig:smoothvsnonsmooth}
\end{figure}

Indeed, it is also known that the $2$-Wasserstein mean of two univariate (continuous) Gaussian densities of mean and standard deviation $(\mu_1,\sigma_1)$ and $(\mu_2,\sigma_2)$ respectively is a Gaussian of mean $(\mu_1+\mu_2)/2$ and standard deviation $(\sigma_1+\sigma_2)/2$~\cite[\S6.3]{Carlier_wasserstein_barycenter}. This fact is illustrated in the top-left plot of Figure~\ref{fig:smoothvsnonsmooth} where we display the average Wasserstein average $\Ncal(0,5/8)$ of the two densities $\Ncal(2,1)$ and $\Ncal(-2,1/4)$. That plot is obtained by using smoothed spline interpolations of a uniformly spaced grid of $100$ values, as can be better observed in the top-right (stair) plot, where the discrete evaluations of these densities are respectively denoted $p_W$, $q_1$ and $q_2$.

Naturally, one would expect the barycenter of $q_1$ and $q_2$ to be close, in some sense, to the discretized histogram $p_W$ of their true barycenter. Histogram $p^\star$, displayed in the bottom-left plot, is the exact optimal solution of Eq.~\eqref{eq-WBP-simplex}, computed with the simplex method. That WBP reduces to a linear program of $2\times 100^2$ variables and $300$ constraints. We observe that $W_2^2(p^\star,q_1)+W_2^2(p^\star,q_2)=0.5833950$ whereas  $W_2^2(p_W,q_1)+W_2^2(p_W,q_2)=0.5834070$. The solution obtained with the simplex has, indeed, a smaller objective than the discretized version of the true barycenter.

The bottom-right plot displays the solution of the \emph{smoothed} Wasserstein barycenter problem (with smoothing parameter $\gamma=\frac{1}{100}$ and a ground cost $M$ that has been re-scaled to have a median value of $1$). The objective value for that smoothed approximation is $0.5834597$.

\begin{figure}[ht]
\centering\includegraphics[width=.75\columnwidth]{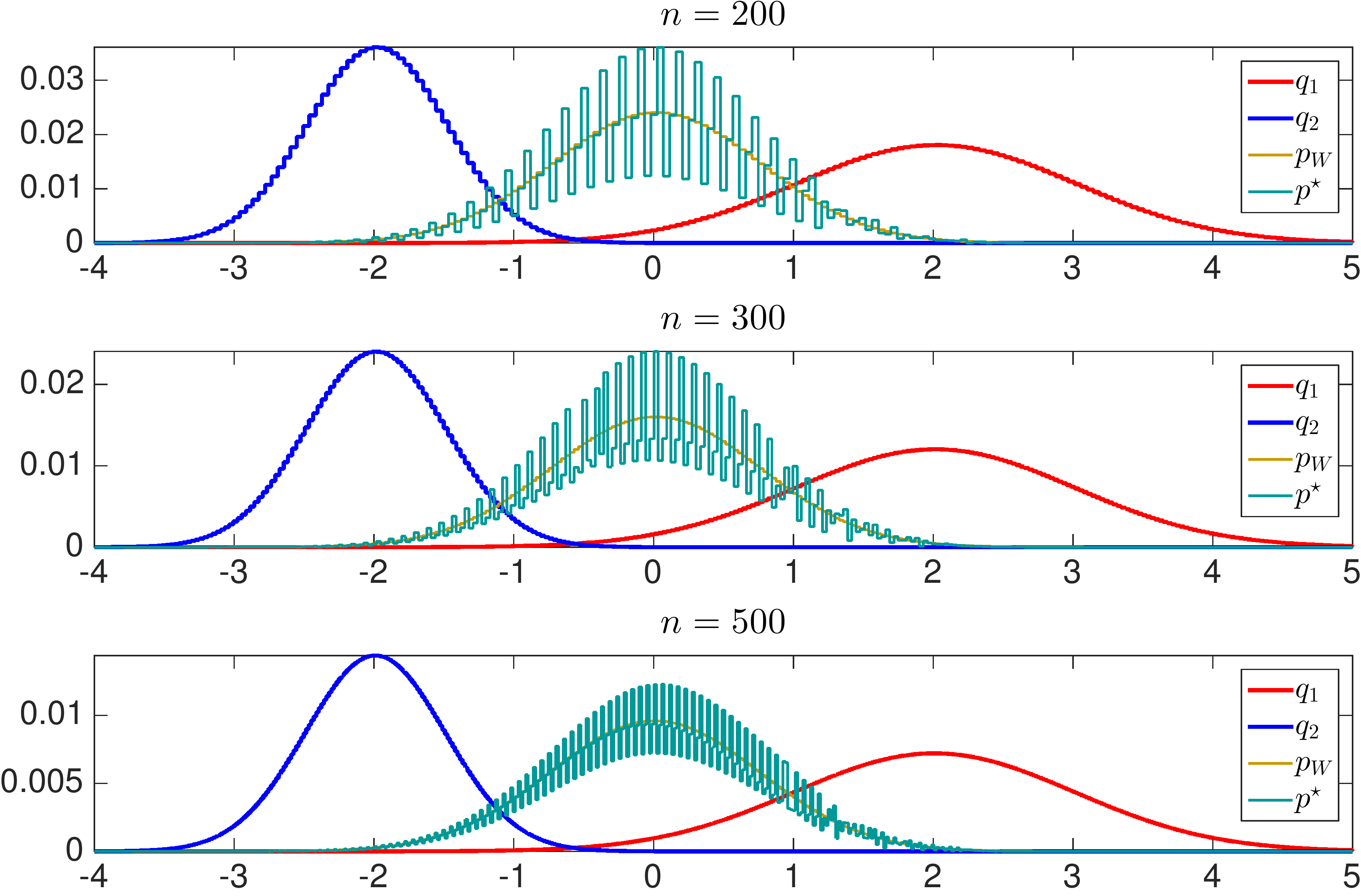}    
\caption{Plots of the exact barycenters for varying grid size $n$.}
\label{fig:smoothvsnonsmooth2}
\end{figure}

This numerical experiment does not contradict the fact that the discretized barycenter $p^\star$ converges to the continuous barycenter as the grid size tends to zero, as shown in~\cite{Carlier-NumericsBarycenters}. This observation illustrates however that, because it is defined as the $\argmin$ of a linear program, the true Wasserstein barycenter may be extremely unstable, even for such a simple problem and for large $n$ as illustrated in Figure~\ref{fig:smoothvsnonsmooth2}. Regularizing the Wasserstein distances has thus the added benefit of smoothing the resulting solution of the Wasserstein barycenter problem, and that of mitigating low sample size effects.

% ------> Version avec une seule figure, mises côte à côte. je trouve que les fonts sont trop petites.
%
% \begin{figure}[ht]
% 	\includegraphics[width=.5\linewidth]{smoothvsnonsmooth.pdf}    
% 	\includegraphics[width=.5\linewidth]{nonsmoothsnot_conv.pdf}
% 	\caption{(top-left) two Gaussian densities and their barycenter (top middle) same densities, discretized (bottom left) discretization of the true barycenter \emph{vs.} the optimum of Eq.~\ref{eq-WBP-simplex} (bottom middle) barycenter computed with our smoothing approach. (right) Plots of the exact barycenters for varying grid size $n$}\label{fig:smoothvsnonsmooth}
% \end{figure}

%%%%%%%%%%%%%%%%%%%%%%%%%%%%%%%%%%%%%%%%%%%%
\subsection{Performance on the Wasserstein Barycenter Problem}
We compare in this section the behavior of the smooth dual approach presented in this paper with that of (i) the smooth primal approach of~\cite{cuturi2014fast}, (ii) the dual approach of \cite{Carlier-NumericsBarycenters}, and (iii) the Bregman iterative projections approach of \cite{DBLP:journals/siamsc/BenamouCCNP15}. We compare these methods on the simple task of computing the Wasserstein barycenter of $12$ histograms laid out on the $100\times100$ grid, as previously introduced in \cite[\S3.2]{DBLP:journals/siamsc/BenamouCCNP15}. We outline briefly all four methods below, and follow by presenting numerical results.

% \begin{figure}[h!]
% \centering\includegraphics[width=\columnwidth]{mymixture.png}
% 	\caption{12 measures, truncated mixtures of Gaussians, used in our benchmark. Convergence speed results displayed in Figure~\ref{fig:costs3} and barycenters obtained in~\ref{fig:bar}}.\label{fig:mix}
% \end{figure}

\begin{figure}[h]
	\centering\includegraphics[width=\columnwidth]{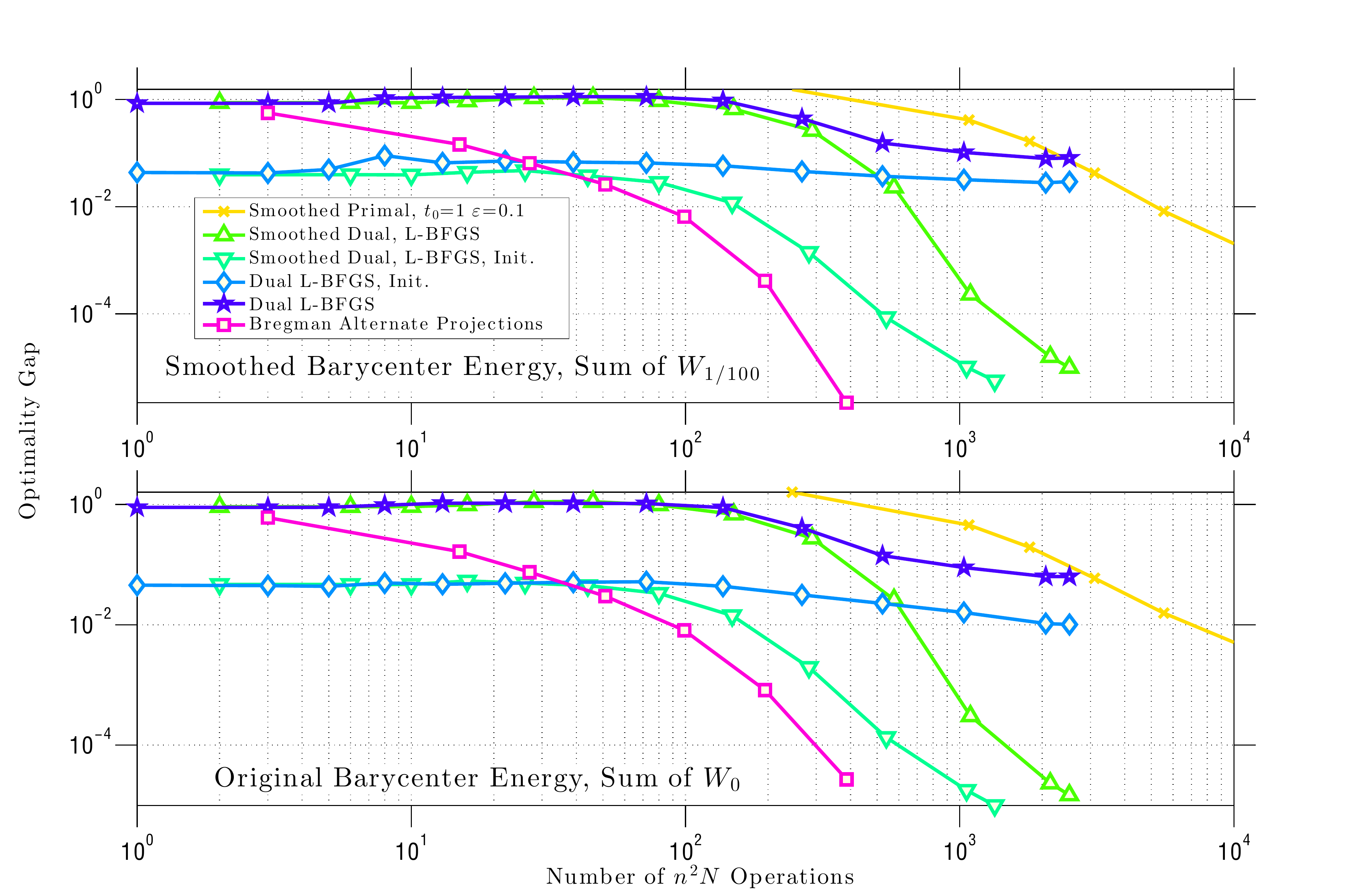}
	\caption{Number of quadratic operations (matrix vector product or min search in a matrix) \emph{vs.} optimization gap to the smallest possible objective found after running $10^5$ iterations of all algorithms, \emph{log-log} scale. Because the smooth primal/dual approaches optimize a \emph{different} criterion than the dual approach, we plot \emph{both} objectives.  The Smooth dual L-BFGS converges faster in \emph{both} smooth and non-smooth metrics. Note the crucial importance of the initialization proposed in \S\ref{sec:magictrick}.}
\end{figure}

% \begin{figure*}[ht]
% 	\centering\includegraphics[width=\textwidth]{12barycenters.png}
% 	\caption{Barycenters obtained for the three different techniques using the data described in Figure~\ref{fig:mix} after at most $10^4$ iteration units, each iteration unit being equal to $n^2N$ operations, here $(100\times 100)^2\times 12$.}\label{fig:bar}
% \end{figure*}

\paragraph{Smooth primal first-order descent} \cite[\S5]{cuturi2014fast} proposed to minimize directly Eq.~\eqref{eq-variational-barycenter-discrete} with a regularizer $\ga>0$. That objective can be evaluated by running $N$ Sinkhorn fixed-point iterations in parallel. That objective is differentiable and its gradient is equal to $\ga\sum_k \la_k \log \alpha_k$, where the $\alpha_k$ are the left scalings obtained with that subroutine. A weakness of that approach is that a tolerance $\epsilon$ for the Sinkhorn fixed-point algorithm must be chosen. Convergence for the Sinkhorn algorithm can be measured with a difference in $l_1$ norm (or any other norm) between the row and column marginals of $\diag(\alpha_k)e^{-M/\ga}\diag(\beta_k)$ and the targeted histograms $\p$ and $\q_k$. Setting that tolerance $\epsilon$ to a large value ensures a faster convergence of the subroutine, but would result in noisy gradients which could slow the convergence of the algorithm. Because the smoothed dual approach only relies on closed form expressions we dot not have to take into account such a trade-off.

\paragraph{Iterative Bregman Projections} \cite[Prop. 1]{DBLP:journals/siamsc/BenamouCCNP15} recalls that the computation of the smoothed Wasserstein distance between $p,q$ using the Sinkhorn algorithm can be interpreted as an iterative alternated projection of the $n\times n$ kernel matrix $e^{-M/\ga}$ onto two affine sets, $\{X:X\ones_n=p\}$ and $\{X:X^T\ones_n=q\}$. That projection is understood to be in the Kullack-Leibler divergence sense. More interestingly, the authors also show that the smoothed WBP itself can also be tackled using an iterative alternated projection, cast this time in a space of dimension $ n\times n \times N$. Very much like the original Sinkhorn algorithm, these projections can be computed for a cheap price, by only tracking variables of size $n\times N$. This approach yields an extremely simple, parameter-free, generalization of the Sinkhorn algorithm which can be applied to the WBP.

\paragraph{Smooth dual L-BFGS} 
The dual formulation with variables $(g_1,\cdots,g_N)\in(\RR^n)^N$ of Eq.~\eqref{eq-dual-pbm} can be solved using a constrained L-BFGS solver
At each iteration of that minimization, we can recover a feasible solution $\p$ to the primal problem of Eq.~\eqref{eq-variational-barycenter-discrete} via the primal-dual relation
$\p = \frac{1}{N}\sum_k \nabla H_{\q_k}^*(\tilde g_k).$

\paragraph{Dual $(\ga=0)$ with L-BFGS} This approach amounts to solving directly the (non-differentiable) dual problem described in Eq.~\eqref{eq-dual-pbm} with no regularization, namely $\ga=0$. Subgradients for the Fenchel-Legendre transforms $H_{q_k}^*$ can be obtained in closed form through Proposition~\ref{prop:harddual}. As with the smoothed-dual formulation, we can also obtain a feasible primal solution by averaging subgradients. We follow \cite{Carlier-NumericsBarycenters}'s recommendation to use L-BFGS. The non-smoothness of that energy is challenging: We have observed empirically that a naive subgradient method applied to that problem fails to converge in all examples we have considered, whereas the L-BFGS approach converges, albeit without guarantees.

%, each a mixture of truncated Gaussians on the $100\times 100$ grid, considered to compute the barycenters displayed in Figure
\paragraph{Averaging Truncated Mixtures of Gaussians}
We consider the 12 truncated mixtures of Gaussians introduced in \cite[\S3.2]{DBLP:journals/siamsc/BenamouCCNP15}. To compare computational time, we use $Nn^2$ elementary operations as the computation unit. These $Nn^2$ operations correspond to matrix-matrix products in the smoothed Wasserstein case, and $Nn$ computations of nearest neighbor assignments among $n$ possible neighbors. Note that in both cases (Gaussian matrix product and nearest neighbors under the $L_2$ metric) computations can be accelerated by considering fast Gaussian convolutions and $kd$-trees for fast nearest neighbor search. We do not consider them in this section. We plot the optimality gap w.r.t the optimum of these 4 techniques as a function of the number of computations, by taking as a reference the lowest value attained across all methods. This value is attained, as in \cite{DBLP:journals/siamsc/BenamouCCNP15}, by the iterative Bregman projections approach after 771 iterations (not displayed in our graph). We show these gaps for both the smoothed $(\gamma=1/100)$ and non-smoothed objectives $(\gamma=0)$. We observe that the iterative Bregman approach outperforms all other techniques. The smoothed-dual approach follows closely, notably when initialized with the formula provided in Definition~\ref{def:SI}.

% We also plot the solutions of all 3 algorithms in Figure~\ref{fig:bar} after up to $10^4$ iterations. Because the smoothed-primal and our smoothed-dual approach aim at minimizing the same objective, it is not surprising that their solutions are similar. Note however that with a budget of at most $10^4$ iterations the solution obtained with dual smoothing is more detailed than the one obtained with a primal descent. The right-most figure, obtained following \cite{Carlier-NumericsBarycenters}'s approach, shows that the original Wasserstein barycenter problem formulation, without smoothing, can yield very irregular solutions, as was also observed by the authors themselves in their paper.

% !TEX root = ../WassersteinDual.tex

%%%%%%%%%%%%%%%%
\section{Regularized Problems}
\label{sec:exten}

We show in this section that our dual optimization framework is versatile enough to deal with functionals involving Wasserstein distances that are more general than the initial WBP problem.

%%%%%%%%%%%%%%%%%%%%%%%%%%%%%%%%%%%%%%%%%%%%%%%%%%%%
\subsection{Regularized Wasserstein Barycenters}\label{sec:regularized}

In order to enforce additional properties of the barycenters, it is possible to penalize~\eqref{eq-variational-barycenter-discrete} with an additional convex regularization, and consider
\eql{\label{eq-regul-primal}
	\umin{p} \sum_{k=1}^N w_k H_{q_k}(p) + J(\Aa p),
}
where $J$ is a convex real-valued function, and $\Aa$ is a linear operator.

The following proposition shows how to compute such a regularized barycenter through a dual optimization problem. 

\begin{prop}
The dual problem to~\eqref{eq-regul-primal} reads
\eql{\label{eq-regul-dual}
	\umin{(u_k)_{k=1}^N, v} \sum_{k=1}^N w_i H_{q_k}^*(u_k) + J^*(v) + \iota_{H}((u_k)_k,v)
}
\eq{
	\qwhereq
	H \eqdef \enscond{ ((u_k)_{k=1}^N, v) }{ \Aa^* v + \sum_k w_k u_k = 0 },
}
and the primal-dual relationships read
\eql{\label{eq-primal-dual}
	\foralls k=1,\ldots,N, \quad p = \nabla H_{q_k}^*(u_k). 
}
\end{prop}

\begin{proof}
	We re-write the initial program~\eqref{eq-regul-primal} as 
	\eql{\label{eq-primal-reformulated}
		\umin{\pi} F( B\pi ) + G(\pi)
	} 
	where we denoted, for $\pi=(p,p_1,\ldots,p_N)$, 
	\begin{align*}
			B \pi &\eqdef (\Aa p, p, p_1,\ldots,p_N) \\
			F(\beta,q, p_1,\ldots,p_N) &\eqdef J(\beta) + \iota_{C}(q,p_1,\ldots,p_N),\\
			G(p, p_1,\ldots,p_N) &\eqdef \sum_k w_k H_{q_k}(p_k)
	\end{align*} 
	for $C \eqdef \enscond{(q,p_1,\ldots,p_N)}{\forall k, p_k=q}$.
	The Fenchel-Rockafelear dual to~\eqref{eq-primal-reformulated} reads
	\eq{
		\umax{\nu = \{v,u,(u_k)_k\}} - F^*(\nu) - G^*(-B^* \nu) 
	}
	where 
	\begin{align*}
		G^*(u,u_1,\ldots,u_N) &= \sum_k w_k H_{q_k}^*(u_k/w_k) + \iota_{\{0\}}( u ),  \\
		B^*(\nu) &= (\Aa^* v + u, u_1,\ldots,u_N), \\
		F^*(\nu) &= J^*(v) + \iota_{C^\bot}(u,u_1,\ldots,u_N), 
	\end{align*} 
	where $C^\bot = \enscond{ (u,u_1,\ldots,u_N) }{u + \sum_k u_k=0}$. 
	One thus obtains the dual
	\eq{
		\umin{ v,b,(u_k)_k } \sum_k w_k H_{q_k}^*(-u_k/w_k)
		\qstq
		\choice{
				\Aa^* v + u = 0, \\
				u + \sum_k u_k=0.
		}
	}	
	The primal-dual relationships reads $\pi \in \partial G^*(-B^* \nu)$, and hence~\eqref{eq-primal-dual}. 
	Changing $-u_k/w_k$ into $u_k$ give the desired formula. 
\end{proof}

Relevant examples of penalizations $J$ include:
\begin{itemize}
	\item In order to enforce some spread of the barycenter, one can use $\Aa=\Id$ and $J(p) = \frac{\la}{2}\norm{p}^2$, in which case $J^*(g) = \frac{1}{2 \la}\norm{g}^2$. In contrast to~\eqref{eq-dual-pbm}, the dual problem~\eqref{eq-regul-dual} is equivalent to an unconstraint smooth optimization. This problem can be solved using a simple Newton descent.  
	\item One can also enforce that the barycenter entries are smaller than some maximum value $\rho$ by setting $\Aa=\Id$ and $J = \iota_{\Cc}$ where $\Cc = \enscond{p}{\normi{p} \leq \rho}$. In this case, one has $J^*(g) = \rho \norm{g}_1$. The optimization~\eqref{eq-regul-dual} is equivalent an unconstrained non-smooth optimization. Since the penalization is an $\ell^1$ norm, one solve it using first order proximal methods as detailed in Section~\ref{sec-first-order-split} bellow.
	\item To force the barycenter to assume some fixed values $p_I^0 \in \RR^{|I|}$ on a given set $I$ of indices, one can use $\Aa=\Id$ and $J=\iota_{\Cc}$ where $\Cc = \enscond{p}{p_I = p_I^0}$ where $p_I=(p_i)_{i \in I}$. One then has $J^*(g) = \dotp{g_I}{p_I^0} + \iota_{\{0\}}(g_{I^c})$.
	\item To force the barycenter to have some smoothness, one can select $\Aa$ to be a spacial derivative operator (for instance a gradient approximated on some grid or mesh) and $J$ to be a norm such as an $\ell^2$ norm (to ensure uniform smoothness) or an $\ell^1$ norm (to ensure piecewise regularity). We explore this idea in Section~\ref{sec-tv-regul}.  
\end{itemize}

%%%%%%%%%%%%%%%%%%%%%%%%%%%%%%%%%%%%%%%%%%%%%%%%%%%%
\subsection{Resolution using First Order Proximal Splitting}
\label{sec-first-order-split}

Assuming without loss of generality that $w_N \neq 0$ (otherwise one simply needs to permute the ordering of the input densities), one can note that it is possible to remove $u_N$ from~\eqref{eq-regul-dual} by imposing, for $x = ((u_k)_{k=1}^{N-1}, v)$
\eq{
	u_N(x) \eqdef - \frac{\Aa^*v}{w_N} - \sum_{i=1}^{N-1} \frac{w_k}{w_N}  u_k, 
} 
and then one can consider the following optimization problem without the $H$ constraint
\eql{\label{eq-primal-unconstr}
	\umin{x} F(x) + G(x)
	\qwhereq
	\choice{
	F(x) \eqdef \sum_{k=1}^{N-1} w_i H_{q_k}^*(u_k) 
		+ w_N H_{q_N}^*\pa{  u_N(x) } \\
	G(x) \eqdef J^*(v).
	}
}

We assume that one is able to compute the proximal operator of $J^*$
\eql{\label{eq-proximal-map}
	\Prox_{\tau J^*}(v) \eqdef \uargmin{v'} \frac{1}{2}\norm{v-v'}^2 + \tau J^*(v').
}
It is for instance an orthogonal projector on a convex set $C$ when $J^*=\iota_{C}$ is the indicator of $C$. 
One can compute easily this projection for instance when $J$ is the $\ell^2$ or the $\ell^1$ norm (see Section~\ref{sec-tv-regul}). 
We refer to~\cite{BauschkeCombettes11} for more background on proximal operators.

The proximal operator of $G$ is then simply
\eq{
	\foralls x=((u_k)_{k=1}^{N-1}, v), \quad \Prox_{\tau G}(x) = ( (u_k)_{k=1}^{N-1}, \Prox_{\tau J^*}(v) ). 
}
Note also that the function $F$ is smooth with a Lipschitz gradient, and that
\eq{
	\nabla F( (u_k)_{k=1}^{N-1}, v ) = 
	\pa{  
		(
			w_k ( \nabla H_{q_k}^*(u_k) 
			-
			\nabla H_{q_N}^*(u_N)  )
		)_{k=1}^{N-1}, 		
		-\Aa \nabla H_{q_N}^*(u_N)
	}
}

The simplest algorithm to solve~\eqref{eq-primal-unconstr} is the Forward-Backward algorithm, whose iteration read
\eql{\label{eq-algo-fb}
	\IIT{x} = \Prox_{\tau J^*}\pa{ \IT{x} - \tau \nabla F(\IT{x}) }.
}
It $\tau<2/L$ where $L$ is the Lipschitz constant of $\nabla F$, then $\IT{x}$ converge to a solution of~\eqref{eq-primal-unconstr}, see~\cite{BauschkeCombettes11} and the references therein. In order to accelerate the convergence of the method, one can use accelerated schemes such as FISTA's algorithm~\cite{beck-fista}.

%%%%%%%%%%%%%%%%%%%%%%%%%%%%%%%%%%%%%%%%%%%%%%%%%%%
\subsection{Total Variation Regularization}
\label{sec-tv-regul}

A typical example of regularization to enforce some geometrical regularity in the barycenter is the total variation regularization on a grid in $\RR^d$ (e.g. $d=2$ for images). It is obtained by considering
\eql{\label{eq-exmp-tv}
	\Aa p \eqdef \nabla p = ( \nabla_i p )_i
	\qandq
	J(u) \eqdef \la \sum_i \norm{u_i}_{\beta}, 
}
where $\nabla_i p \in \RR^d$ is a finite difference approximation of the gradient at a point indexed by $i$, and $\la \geq 0$ is the regularization strength.
When using the $\ell^2$ norm to measure the gradient amplitude, i.e. $\beta=2$, one obtains the so-called isotropic total variation, that tends to round corners, and essentially penalizes the length of the level sets of the barycenter, possibly merging clusters together. 
When using instead the $\ell^1$ norm, i.e. $\beta=1$, one obtains the so-called anisotropic total variation, which penalizes independently horizontal and vertical derivative, thus favoring the emergence of axis-aligned edges, and giving a ``crystalline'' look to the barycenters. We refer for instance to~\cite{2015-Caselles-tv} for a study of the effect of TV regularization on the shapes of levelsets using isotropic and crystalline total variations. 

In this case, it is possible to compute in closed form the proximal operator~\eqref{eq-proximal-map}. Indeed, one has $J^* = \iota_{\norm{\cdot}_{\beta^*} \leq \la}$ where $\beta^*$ is the conjugate exponent $1/\beta+1/\beta^*=1$. One can compute explicitly the proximal operator in the case $\beta \in \{1,2\}$ since they correspond to orthogonal projectors on $\ell^{\beta^*}$ balls
\eq{
	\Prox_{\tau J^*}(v)_i = 
	\choice{
		\min(\max(v_i,-\la), \la) \qifq \beta=1,  \\
		v_i \frac{\la}{ \max(\norm{v_i},1) }  \qifq \beta=2.
	}
}

%%%%%%%%%%%%%%%%%%%%%%%%%%%%%%%%%%%%%%%%%%%%%%%%%%%
\subsection{Barycenters of Images}
\label{sec-tv-images}

We start by computing barycenters of a small number of 2-D images, that are discretized on an uniform rectangular grid of $n=256 \times 256$ pixels $(z_i)_{i=1}^N$. We use either the isotropic ($\beta=2$) or anisotropic ($\beta=1$) total variation presented above, where $\nabla$ is defined using standard forward finite differences along each axis, and using Neumann boundary conditions. The metric is the usual squared Euclidean metric
\eql{\label{eq-squared-eucl-metric}
	\foralls (i,j) \in \{1,\ldots,n\}^2, \quad
	M_{i,j} = \norm{z_i-z_j}^2. 
}
The Gibbs kernel $K=e^{-M/\ga}$ is a filtering with a Gaussian kernel, that can be applied efficiently to histograms in nearly linear time, see~\cite{2015-solomon-siggraph} for more details about convolutional kernels.

\newcommand{\myPic}[3]{\includegraphics[width=.19\linewidth]{tv-regularization/bary4-#1/lambda#2/bary#3}}
\newcommand{\myspace}{\hspace{0mm}}

\newcommand{\myTable}[3]{ % 
	\begin{subfigure}[b]{0.48\textwidth}
	\begin{tabular}{|@{}c@{\myspace{}}c@{\myspace{}}c@{\myspace{}}c@{\myspace{}}c@{}|}
		\hline
		\myPic{#1}{#2}{1}&
		\myPic{#1}{#2}{2}&
		\myPic{#1}{#2}{3}&
		\myPic{#1}{#2}{4}&
		\myPic{#1}{#2}{5}\\[-2mm]
		\myPic{#1}{#2}{6}&
		\myPic{#1}{#2}{7}&
		\myPic{#1}{#2}{8}&
		\myPic{#1}{#2}{9}&
		\myPic{#1}{#2}{10}\\[-2mm]
		\myPic{#1}{#2}{11}&
		\myPic{#1}{#2}{12}&
		\myPic{#1}{#2}{13}&
		\myPic{#1}{#2}{14}&
		\myPic{#1}{#2}{15}\\[-2mm]
		\myPic{#1}{#2}{16}&
		\myPic{#1}{#2}{17}&
		\myPic{#1}{#2}{18}&
		\myPic{#1}{#2}{19}&
		\myPic{#1}{#2}{20}\\[-2mm]
		\myPic{#1}{#2}{21}&
		\myPic{#1}{#2}{22}&
		\myPic{#1}{#2}{23}&
		\myPic{#1}{#2}{24}&
		\myPic{#1}{#2}{25}\\\hline
	\end{tabular}
	\caption{#3}
	\end{subfigure}}

\begin{figure}[h!]
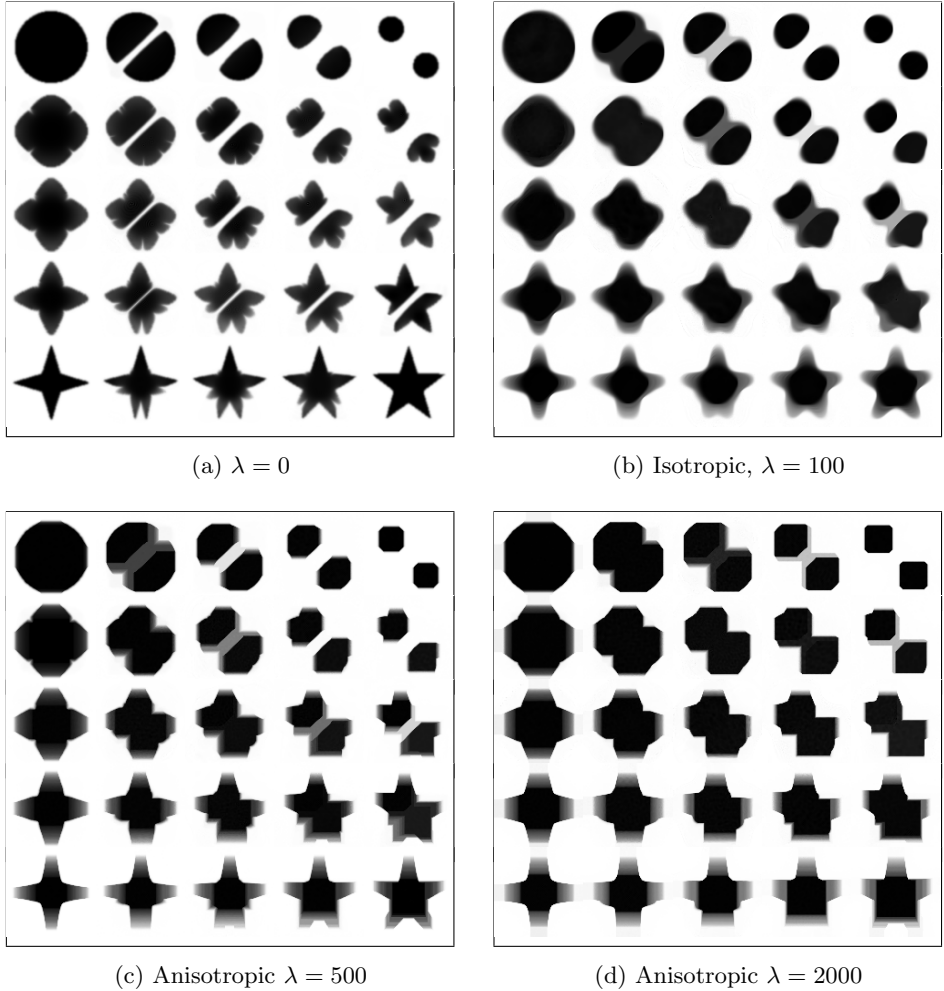

	\centering	
	\myTable{aniso}{0}{$\la=0$}	 
	\myTable{iso}{1000}{Isotropic, $\la=100$}	 
	\myTable{aniso}{500}{Anisotropic $\la=500$}	 
	\myTable{aniso}{2000}{Anisotropic $\la=2000$}	 
	\caption{% 
		Examples of isotropic and anisotropic TV regularization for the computation of barycenters between four input densities. 
		The weights $(w_k)_{k=1}^N$ are bilinear interpolation weights, so that it is for instance $w=(1,0,0,0)$ on the 
		top left corner and $(0,0,0,1)$ on the bottom right corner.  \label{fig-images-bary}
	}
\end{figure}

Figure~\ref{fig-images-bary} shows examples of barycenters of $N=4$ input histograms computed by solving~\eqref{eq-regul-primal} using the projected gradient descent method~\eqref{eq-algo-fb}. The input histograms represent 2-D shapes, and are uniform (constant) distributions inside the support of the shapes. Note that in general the barycenters are not shapes, i.e. they are not uniform distributions, but this method can nevertheless be used to define meaningful averaging of shapes as exposed in~\cite{2015-solomon-siggraph}. 
Figure~\ref{fig-images-bary} compares the effects of $\beta \in \{1,2\}$, and one can clearly see how the isotropic total variation ($\beta=2$) rounds the corners of the input densities, while the anisotropic version ($\beta=1$) favors horizontal/vertical edges.

\newcommand{\myPicL}[2]{\includegraphics[width=.125\linewidth]{tv-regularization/bary2-#1/bary1-lambda#2}}

\begin{figure}[h!]
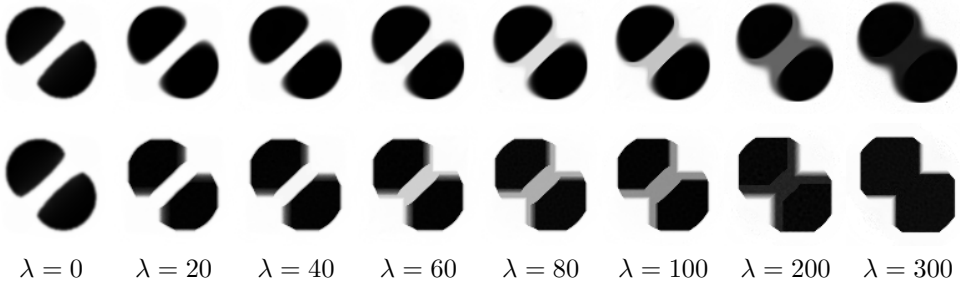

	\centering	
	\begin{tabular}{@{}c@{}c@{}c@{}c@{}c@{}c@{}c@{}c@{}}
		\myPicL{iso}{0} &
		\myPicL{iso}{200} &
		\myPicL{iso}{400} &
		\myPicL{iso}{600} &
		\myPicL{iso}{800} &
		\myPicL{iso}{1000} &
		\myPicL{iso}{2000} &
		\myPicL{iso}{3000} \\
		\myPicL{aniso}{0} &
		\myPicL{aniso}{200} &
		\myPicL{aniso}{400} &
		\myPicL{aniso}{600} &
		\myPicL{aniso}{800} &
		\myPicL{aniso}{1000} &
		\myPicL{aniso}{2000} &
		\myPicL{aniso}{3000} \\
		$\la=0$ &
		$\la=20$ &
		$\la=40$ &
		$\la=60$ &
		$\la=80$ &
		$\la=100$ &
		$\la=200$ &
		$\la=300$ 
	\end{tabular}	 
	\caption{% 
		Influence of $\la$ parameter for the iso-barycenter (i.e. $w=(1/2,1/2)$) between two input densities (they are the upper-left and upper-right corner of the $\la=0$ case in Figure~\ref{fig-images-bary}). 
		\textit{Top row:} isotropic total variation ($\beta=2$).
		\textit{Bottom row:} anisotropic total variation ($\beta=1$). \label{fig-images-influ-lambda}
	}
\end{figure}

Figure~\ref{fig-images-influ-lambda} shows the influence of the regularization strength $\la$ to compute the iso-barycenter of $N=2$ shapes. This highlights the fact that this total variation regularization has the tendency to group together small clusters, which might be beneficial for some applications, as illustrated in Section~\ref{sec-tv-meg} on MEG data denoising.

%%%%%%%%%%%%%%%%%%%%%%%%%%%%%%%%%%%%%%%%%%%%%%%%%%%
\subsection{Barycenters of MEG Data}
\label{sec-tv-meg}

We applied our method to a magnetoencephalography (MEG) dataset. In this setup, brain activity of a subject is recorded  (Elekta Neuromag, 306 sensors of which 204 planar gradiometers and 102 magnetometers, sampling frequency 1000Hz) while the subject reacted to the presentation of a target stimulus by pressing either the left or the right button.

Data is preprocessed applying signal space separation correction, interpolation of noisy sensors, and realignment of data into a subject-specific head position (MaxFilter, Elekta Neuromag). The signal was then filtered (low pass 40HZ), and artifacts such as blinks and heartbeats removed thanks to Signal-Space Projection using the Brainstorm software\footnote{\url{http://neuroimage.usc.edu/brainstorm}}. The samples we used for our barycenter computations are an average of the norm of the two gradiometers for each channel from stimulation onto 50ms and the classes were left or right button. 

\newcommand{\myPicMEGin}[2]{\includegraphics[width=.125\linewidth]{meg/input/#1-#2}}
\newcommand{\myPicMEGbar}[2]{\includegraphics[width=.125\linewidth]{meg/bary/bary#1-lambda#2}}
\newcommand{\myPicMEGmean}[1]{\includegraphics[width=.125\linewidth]{meg/bary/bary#1-mean}}

\begin{figure}[h!]
	\centering	
	\begin{tabular}{@{}c@{}c@{}c@{}c@{}|@{}c@{}c@{}c@{}c@{}}
		& Class 1 & & & %
		& Class 2 & & \\ % 
		\myPicMEGin{1}{1} &
		\myPicMEGin{1}{2} &
		\myPicMEGin{1}{3} &
		\myPicMEGmean{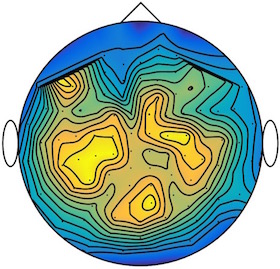} & 
		%%%%
		\myPicMEGin{2}{1} &
		\myPicMEGin{2}{2} &
		\myPicMEGin{2}{3} &
		\myPicMEGmean{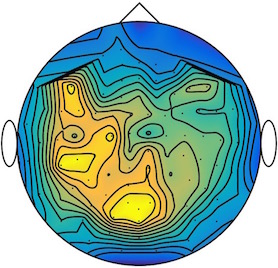} \\
		Sample 1 & 
		Sample 2 &
		Sample 3 &
		Mean &
		Sample 1 & 
		Sample 2 &
		Sample 3 &
		Mean  \\
		\myPicMEGbar{1}{0} &
		\myPicMEGbar{1}{2} &
		\myPicMEGbar{1}{4} &
		\myPicMEGbar{1}{8} &
		%%%
		\myPicMEGbar{2}{0} &
		\myPicMEGbar{2}{2} &
		\myPicMEGbar{2}{4} &
		\myPicMEGbar{2}{8} \\
		$\la=0$ &
		$\la=2$ &
		$\la=4$ &
		$\la=8$ &		
		%%%
		$\la=0$ &
		$\la=2$ &
		$\la=4$ &
		$\la=8$ 
	\end{tabular}	 
	\caption{% 
		Barycenter computation on MEG data. The left/right panels shows respectively the first and the second class, 
		corresponding to recordings where the subject is asked to push the left or the right button.
		\textit{Top row:} examples of input histograms $q_k$ for each class, as well as the 
		$\ell^2$ mean $N^{-1} \sum_k q_k$.
		\textit{Bottom row:} computed TV-regularized barycenter for different values of $\la$
		($\la=0$ corresponding to no regularization). \label{fig-meg}
	}
\end{figure}

This results in two classes of recordings, one for each pressed button. We aim at computing a representative activity map for each class using Wasserstein barycenters. For each class we have $N=33$ recordings $(q_k)_{k=1}^N$ each having $n=66$ samples located on the vertices of an hexahedral mesh of a hemisphere (corresponding to a MEG recording helmet). These recorded values are positive by construction, and we rescale them linearly to impose $q_k \in \Si_n$.  Figure~\ref{fig-meg}, top row, shows some samples from this dataset, displayed using interpolated colors as well as iso-level curves. The black dots represent the position $(z_i)_{i=1}^n$ of the electrodes on the half-sphere of the helmet, flattened on a 2-D disk. 

We computed TV-regularized barycenters independently for each class by solving~\eqref{eq-regul-primal} with the TV regularization using the projected gradient descent method~\eqref{eq-algo-fb}. We used a squared Euclidean metric~\eqref{eq-squared-eucl-metric} on the flattened hemisphere.
Since the data is defined on an irregular graph, instead of~\eqref{eq-exmp-tv}, we use a graph-based discrete gradient. We denote $( (i,j) )_{(i,j) \in \Gg}$ the graph which connects neighboring electrodes. The gradient operator on the graph is
\eq{
	\foralls p \in \RR^n, \quad
	\Aa p \eqdef ( p_i - p_j )_{ (i,j) \in \Gg } \in \RR^{|\Gg|}.
}
The total variation on this graph is then obtained by using $J = \la \norm{\cdot}_1$, the $\ell^1$ norm, i.e. we use $\beta=1$ in~\eqref{eq-exmp-tv}. 

Figure~\ref{fig-meg} compares the naive $\ell^2$ barycenters (i.e. the usual mean), barycenters obtained without regularization (i.e. $\la=0$) and barycenters computed with an increasing regularization strength $\la$. The input histograms $(p_k)_k$ being very noisy, the use of regularization is important to make the area of significant activity emerge from the noise. The use of a TV regularization helps to keep a sharp transition between active and non-active regions.

%%%%%%%%%%%%%%%%%%%%%%%%%%%%%%%%%%%%%%%%%%%%%%%%%%%%
\subsection{Gradient Flow}
\label{sec-grad-flows}

Instead of computing barycenters, we now use our regularization to define time-evolutions, which are defined through a so-called discrete gradient flow. 

Starting from an initial histogram $p_0 \in \Si_n$, we define iteratively 
\eql{\label{eq-gradflow-def}
	p_{k+1} \eqdef \uargmin{p \in \Si_N} H_{p_k}(p) + \tau f(p).
}
This means that one seeks a new iterate at (discrete time) $k+1$ that is both close (according to the Wasserstein distance) to $p_k$ and minimizes the functional $f$. In the following, we consider the gradient flow of regularization functionals as considered before, i.e. that are of the form $\tau f = J \circ \Aa$. Problem~\eqref{eq-gradflow-def} is thus a special case of~\eqref{eq-regul-primal} with $N=1$. 

Letting $k \rightarrow +\infty$, one can informally think of $p_k$ as a discretization of a time evolution evaluated at time $t=k\tau$. This method is a general scheme presented in much detail in the monograph~\cite{ambrosio2006gradient}. The use of an implicit time-stepping~\eqref{eq-gradflow-def} allows one to define time evolutions to minimize functionals that are not necessarily smooth, and this is exactly the case of the total variation semi-norm (since $J$ is not differentiable). 
The use of gradient flows in the context of the Wasserstein fidelity to the previous iterate has been introduced initially in the seminal paper~\cite{jordan1998variational}. When $f$ is the entropy functional, this paper proves that the countinous flow defined by the limit $k \rightarrow+\infty$ and $\tau \rightarrow 0$ is a heat equation. Numerous theoretical papers have shown how to recover many existing non-linear PDE's by considering the appropriate functional $f$, see for instance~\cite{otto2001geometry,GianazzaARMA}.

The numerical method we consider in this article is the one introduced in~\cite{Peyre-JKO}, that makes use of the entropic smoothing of the Wasserstein distance. It is not the scope of the present paper to discuss the problem of approximating gradient flows and the underlying limit non-linear PDE's, and we refer to~\cite{Peyre-JKO} for an overview of the vast literature on this topic. A major bottleneck of the algorithm developed in~\cite{Peyre-JKO} is that it uses a primal optimization scheme (Dykstra's algorithm) that necessitates the computation of the proximal operator of $f$ according to the Kulback-Leibler divergence. Only relatively simple functionals (basically separable functionals such as the entropy) can thus be treated by this approach. In contrast, our dual method can cope with a much larger set of functions, and in particular those of the form $f = J \circ \Aa$, i.e. obtained by pre-composition with a linear operator.

\newcommand{\myPicGF}[2]{\includegraphics[width=.16\linewidth]{gradient-flow/#1/#1-flow-#2}}

\begin{figure}[h!]
	\centering	
	\begin{tabular}{@{}c@{\hspace{1mm}}c@{\hspace{1mm}}c@{\hspace{1mm}}c@{\hspace{1mm}}c@{\hspace{1mm}}c@{}}
		\myPicGF{randdisks}{0} &
		\myPicGF{randdisks}{2} &
		\myPicGF{randdisks}{5} &
		\myPicGF{randdisks}{10} &
		\myPicGF{randdisks}{15} &
		\myPicGF{randdisks}{20}  \\
		\myPicGF{hibiscus}{0} &
		\myPicGF{hibiscus}{2} &
		\myPicGF{hibiscus}{5} &
		\myPicGF{hibiscus}{10} &
		\myPicGF{hibiscus}{15} &
		\myPicGF{hibiscus}{20}  \\
		$t=0$ &
		$t=20$ &
		$t=40$ &
		$t=60$ &
		$t=80$ &
		$t=100$ 
	\end{tabular}	 
	\caption{% 
		Examples of gradient flows~\eqref{eq-gradflow-def} at various times $t \eqdef k\tau$. \label{fig-gradient-flow}
	}
\end{figure}

Figure~\ref{fig-gradient-flow} shows examples of gradient flows computed for the isotropic total variation $f(p)=\norm{\nabla p}_1$ as defined in~\eqref{eq-exmp-tv}. We use the discretization setup considered in Section~\ref{sec-tv-images}. This is exactly the regularization flow considered by~\cite{Burger-JKO}. This paper defines formally the highly non-linear fourth order PDE corresponding to the limit flow. This is however not a ``true'' PDE since the initial TV functional is non-smooth, and derivatives should be understood in a weak sense, as limit of an implicit discrete time stepping. While the algorithm proposed in~\cite{Burger-JKO} uses the usual (unregularized) Wasserstein distance, the use of a regularized transport allows us to deal with problems of larger sizes, with a faster numerical scheme. The price to pay is an additional blurring introduced by the  entropic smoothing, but this is acceptable for applications to denoising in imaging. Figure~\ref{fig-gradient-flow} illustrates the behavior of this TV regularization flow, which has the tendency to group together clusters of mass, and performs some kind of progressive ``percolation'' over the whole image.

% !TEX root = ../WassersteinDual.tex

%%%%%%%%%%%%%%%%%%%%%%%%%%%%%%%%%%%%%%%%%%%%%%%%%%%
\section*{Conclusion}

In this paper, we introduced a dual framework for the resolution of certain variational problems involving Wasserstein distances. The key contribution is that the dual functional is smooth and that its gradient can be computed in closed form and involves only multiplications with a Gibbs kernel. 
We illustrate this approach with applications to several problems revolving around the idea of Wasserstein barycenters. This method is particularly advantageous for the computation of regularized barycenters, since pre-composition by linear operator  (such as discrete gradient on images or graphs) of functionals is simple to handle. 
Our numerical findings is that entropic smoothing is crucial to stabilize the computation of barycenters and to obtain fast numerical schemes. 
Further regularization using for instance a total variation is also beneficial, and can be used in the framework of gradient flows.

% one that involves computing Wasserstein barycenters and another that involves learning a dictionary and weights with a Wasserstein fit. Our approach has several attractive qualities: \emph{(i)} our entropic regularization ensures the unicity of the optimal solution in the simple WBP problem and facilitates the computation of each of the convex sub-problems considered in dictionary learning; \emph{(ii)} we observe that solutions obtained with this regularization exhibit a level of smoothness which is comparable to that of the original measures. This property can be desirable in some cases. \emph{(iii)} our approach can be initialized very efficiently thanks to a simple rule that is optimal in the simplified case where all original measures are dirac masses. \emph{(iv)} using Fenchel duality, we show that Wasserstein variational problems can be carried out using closed form functions. We believe this class of approaches can be extended to more general tasks and can scale up to more demanding learning problems.

%%%%%%%%%%%%%%%%%%%%%%%%%%%%%%%%%%%%%%%%%%%%%%%%%%%
\section*{Acknowledgments}

The work of Gabriel Peyr\'e has been supported by the European Research Council (ERC project SIGMA-Vision).
Marco Cuturi gratefully acknowledges the support of JSPS young researcher A grant 26700002.
We would like to thank Antoine Rolet, Nicolas Papadakis and Julien Rabin for stimulating discussions. 
We would like to thank Valentina Borghesani, Manuela Piazza et Marco Buiatti for giving us access to the MEG data.
We would like to thank Fabian Pedregosa and the chaire ``\'Economie des Nouvelles Donn\'ees'' for the help in the preparation of the MEG data.

\appendix
% !TEX root = ../WassersteinDual.tex

\section{Legendre Transform with Respect to Two Histograms}\label{sub:two}

Theorem~\ref{thm-legendre-transf} can be extended to study the Legendre transform of $W_\ga(\p,\q)$ with respect to both arguments $(\p,\q)$ instead of only $\p$. Indeed, expression~\eqref{eq-wassdist-dual} shows that $(\p,\q) \mapsto W_\ga(\p,\q)$ is a convex function (as a maximum of linear forms), so that one can define $\foralls (g,h) \in \RR^n \times \RR^n$,
$$	
W^*_\ga(g,h) = \umax{\p,\q \in \Sigma_n} \dotp{g}{\p} + \dotp{h}{\q} - W(\p,\q).
$$

The following proposition adapts  to this setting.

\begin{proposition}\label{eq-obj-bothvar}
	The function $W^*_\ga$ is $C^\infty$ at $(g,h) \in \RR^n \times \RR^n$ and, writing $K=e^{-M/\ga}$, $\alpha=e^{g/\ga}, \beta=e^{h/\ga}$ and $\Kcal_{\alpha\beta} = \diag(\alpha)K \beta$, we have that
			
	\begin{align*}% \label{eq-obj-bothvar}
  		W^*_\ga(g,h) &= -\gamma \log \alpha^TK\beta,\\%\label{eq-grad-bothvar} 
		\nabla W^*_\ga(g,h) &= \frac{1}{\alpha^T K \beta}\begin{bmatrix}\Kcal_{\alpha,\beta} \\ \Kcal_{\beta,\alpha}\end{bmatrix},\\%\label{eq-hessian-bothvar} 
\nabla^2 W_\ga^*(g) &= \frac{1}{\ga\alpha^T K \beta} \begin{bmatrix} A_\gamma(g,h) & B_\gamma(g,h) \\ B_\gamma(h,g ) & A_\gamma(h,g )\end{bmatrix}.				
		\end{align*}
		
	$$	
	\qwhereq					
	\begin{cases}
		A_\gamma(g,h) &=  \diag(\Kcal_{\alpha\beta})- \frac{1}{\alpha^T K \beta}\Kcal_{\alpha\beta}\Kcal_{\alpha\beta}^T,\\
		B_\gamma(g,h) &= \diag(\beta)\K \diag(\alpha)-\frac{1}{\alpha^T K \beta}\Kcal_{\beta\alpha}\Kcal^T_{\alpha\beta}.
	\end{cases}
	$$
	Moreover, the gradient function $(g,h)\mapsto\nabla W^*_\ga(g,h)$ is $2/\gamma$ Lipschitz.	
\end{proposition}	
\begin{proof}
One has that $W^*_\ga(g,h)$ can be written
\begin{align*}
	 &\umax{p,q\in\Si_n} \dotp{g}{p} + \dotp{h}{q} - \umax{u,v} \dotp{u}{p} + \dotp{v}{q} - \beta_{\ga,M}(u,v)  \\
			 &= \umax{p,q}  -\umax{u,v} \dotp{u+g}{p}  + \dotp{v+h}{q} - \beta_{\ga,M}(u,v) \\
			 &= \umax{p,q}  -\umax{u',v'} \dotp{u'}{p} + \dotp{v'}{q} - \beta_{\ga,M}(u'+g,v'+h) \\
%			 &= \umax{p,q}  -\umax{u',v'} \dotp{u'}{p} + \dotp{v'}{q} - \beta_{\ga,M-g\ones^T - \ones h^T}(u',v') \\
			 &= \umax{p,q}  -W_{M+g\ones^T + \ones h^T}(\p,\q) \\
			 &= \umax{p,q}  - \! \umin{X \in U(\p,\q)} \dotp{X}{M-g\ones^T - \ones h^T} - \ga E(X) \\			 
			 &= - \umin{X\in\Sigma_{n^2}} \dotp{X}{M-g\ones^T - \ones h^T} - \ga E(X).
\end{align*}
One verifies that the last Eq. is equivalent to a classic maximal entropy problem which can be solved uniquely with a Gibbs distribution equal to $X^\star$ given below,
%\begin{equation}\label{eq:xgh} 
	$$X^\star = \frac{\diag(\alpha) K \diag(\beta)}{\alpha^T K \beta}.$$%\end{equation}
Substituting this expression in the formula above for $W^*_\ga(g,h)$ yields that
$$
W^*_\ga(g,h) = -\gamma \log \alpha^TK\beta.
$$
Since the gradients with respect to $g$ and $h$ of $W^*_\ga(g,h)$ are $X^\star \ones$ and $X^{\star T}\ones$ respectively, this results in the expression provided above. The Hessian follows from that result, and the Lipschitz continuity of the gradient can be obtained by showing that the Hessian's trace can be upper-bounded by $2/\ga$ by noticing that the trace of both $A_\ga(g,h)$ and $A_\ga(h,g)$ is upper-bounded by $\alpha^TK\beta$.
\end{proof}

\bibliographystyle{plain}
\bibliography{refs}

\end{document}